\documentclass{article}

\usepackage{arxiv}
\usepackage{subfigure}

\usepackage[utf8]{inputenc} % allow utf-8 input
\usepackage[T1]{fontenc}    % use 8-bit T1 fonts
\usepackage{url}            % simple URL typesetting
\usepackage{booktabs}       % professional-quality tables
\usepackage{amsfonts}       % blackboard math symbols
\usepackage{nicefrac}       % compact symbols for 1/2, etc.
\usepackage{lipsum}		% Can be removed after putting your text content
\usepackage{graphicx}
\usepackage{natbib}
\usepackage{doi}
\usepackage{makecell}
\usepackage[table]{xcolor}
\usepackage{longtable}
\usepackage{fancyhdr}

\usepackage{multirow}
\usepackage{caption}
\usepackage{enumitem}
\usepackage{multicol} % 引入多栏包
\usepackage{lipsum} % 用于生成示例文本

\usepackage{array}

\setlist[itemize]{leftmargin=10pt} 
\usepackage{listings}
\usepackage{xcolor}
\definecolor{codegreen}{rgb}{0,0.6,0}
\definecolor{codegray}{rgb}{0.5,0.5,0.5}
\definecolor{codepurple}{rgb}{0.58,0,0.82}
\definecolor{backcolour}{rgb}{0.95,0.95,0.92}

\lstdefinestyle{mystyle}{
    backgroundcolor=\color{backcolour},   
    commentstyle=\color{codegreen},
    keywordstyle=\color{magenta},
    numberstyle=\tiny\color{codegray},
    stringstyle=\color{codepurple},
    basicstyle=\ttfamily\footnotesize,
    breakatwhitespace=false,         
    breaklines=true,                 
    captionpos=b,                    
    keepspaces=true,                 
    numbers=left,                    
    numbersep=5pt,                  
    showspaces=false,                
    showstringspaces=false,
    showtabs=false,                  
    tabsize=2
}

\title{JT-Safe: Intrinsically Enhancing the Safety and Trustworthiness of LLMs}

\author{\textbf{Junlan Feng}\thanks{ Corresponding author:~\texttt{fengjunlan@cmjt.chinamobile.com}}
        ~~~~Fanyu Meng
        ~~~~Chong Long
        ~~~~Pengyu Cong 
        ~~~~Duqing Wang %\\[0.2em]
        ~~~~Yan Zheng \\
        ~~~~Yuyao Zhang
        ~~~~Xuanchang Gao
        ~~~~Ye Yuan
        ~~~~Yunfei Ma  
        ~~~~Zhijie Ren
        ~~~~Fan Yang \\
        ~~~~Na Wu
        ~~~~Di Jin
        ~~~~Chao Deng
        \\[0.3em]
        \textit{China Mobile Jiutian Research}, Beijing, China \\[0.3em]
        \texttt{\{fengjunlan, mengfanyu, longchong, congpengyu, wangduqing, zhengyan,} \\
        \texttt{zhangyuyao, gaoxuanchang, yuanye, mayunfei, renzhijie, yangfan,} \\
        \texttt{wuna, jindi, dengchao\}@cmjt.chinamobile.com}
}
\date{}

\renewcommand{\headeright}{Technical Report}
\renewcommand{\shorttitle}{JT-Safe}
\begin{document}
\maketitle
\vspace{-18pt}

\begin{abstract}
The hallucination and credibility concerns of large language models (LLMs) are global challenges that the industry is collectively addressing. Recently, a significant amount of advances have been made on post-training and inference techniques to mitigate these challenges. However, it is widely agreed that unsafe and hallucinations of LLMs intrinsically originate from pre-training, involving pre-training data and the next-token prediction learning mechanism. In this paper, we focus on enhancing pre-training data to improve the trustworthiness and safety of LLMs. Since the data is vast, it’s almost impossible to entirely purge the data of factual errors, logical inconsistencies, or distributional biases. Moreover, the pre-training data lack grounding in real-world knowledge. Each piece of data is treated as a sequence of tokens rather than as a representation of a part of the world. To overcome these issues, we propose approaches to enhancing our pre-training data with its context in the world and increasing a substantial amount of data reflecting industrial scenarios. We argue that most source data are created by the authors for specific purposes in a certain spatial-temporal context. They have played a role in the real world. By incorporating related world context information, we aim to better anchor pre-training data within real-world scenarios, thereby reducing uncertainty in model training and enhancing the model’s safety and trustworthiness. We refer to our Data with World Context as DWC. We continue pre-training an earlier checkpoint of JT-35B-Base with 1.5 trillion of DWC tokens. We introduce our post-training procedures to activate the potentials of DWC. Compared with the Qwen model of a similar scale, JT-Safe-35B achieves an average performance improvement of 1.79\% on the Safety and Trustworthy evaluation benchmarks, while being pretrained with only 6.2 trillion tokens.

\end{abstract}
\begin{figure}[t]
 \centering
 \includegraphics[width=16cm]{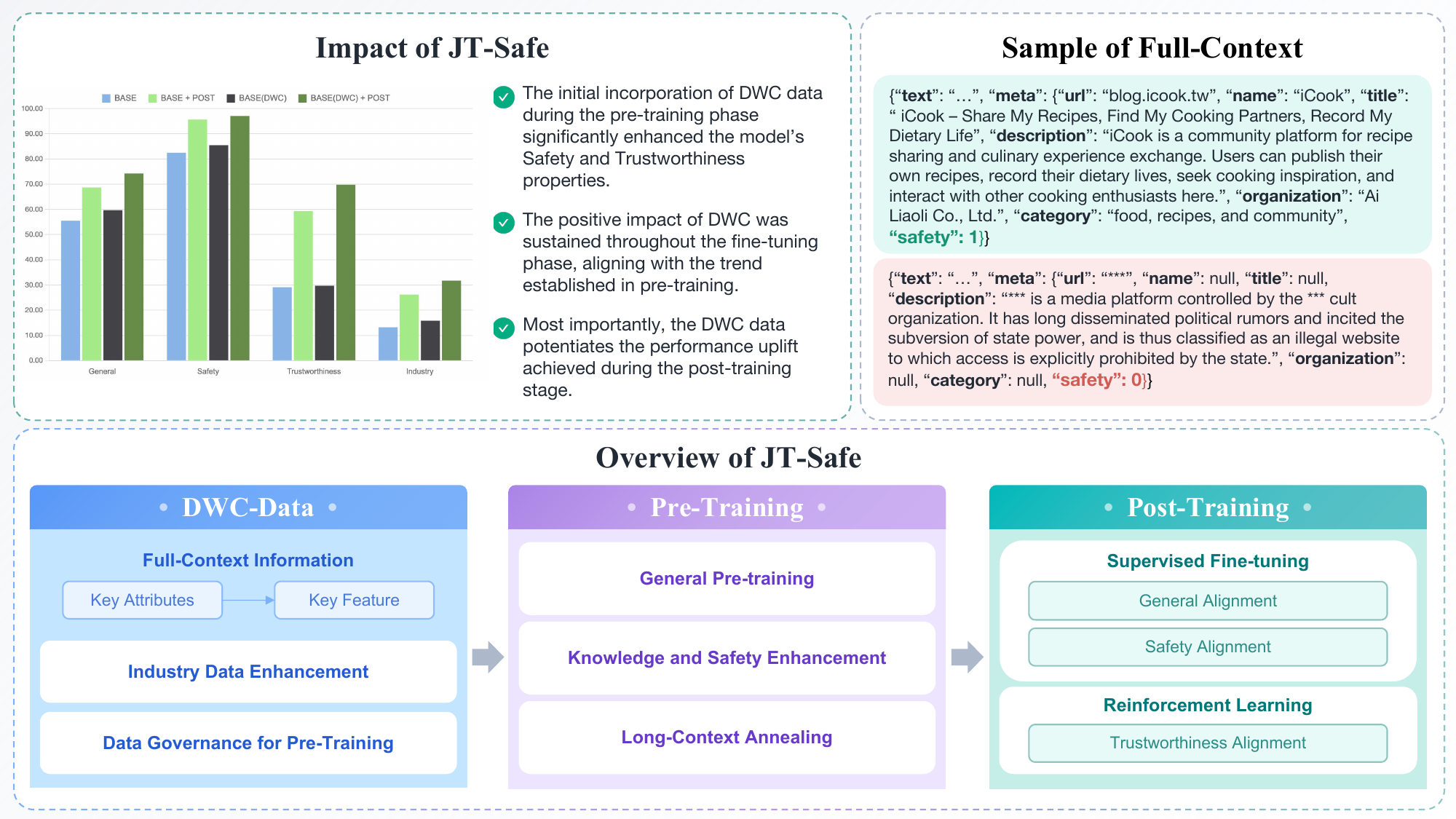}
 \caption{JT-Safe Framework Architecture.}
 \label{fig:intro}
\end{figure}

\section{Introduction}
As the comprehensive capabilities of large language models(LLMs) advance rapidly, their applications have become extremely widespread and diverse. However, language models are known to produce unsafe and overconfident responses~\citep{qwen3guard}~\citep{kalai2025language}. Both open-weight and proprietary models may generate content that is unfair or even harmful to individuals, groups, enterprises, and governments. When facing attacks such as prompt injection, jail-breaking, and multi-modal adversarial examples, these models frequently encounter issues, including security alignment failures, training data leakage, malicious code generation, and harmful content output~\citep{bai2022constitutionalaiharmlessnessai}. Problems of insufficient consistency and trustworthiness in model-generated content often arise in practical applications. To a large extent, these challenges restrict the large-scale deployment of large models and their in-depth application in core production and commercial scenarios.

To address the issue of insufficient security, several models designed for security protection have been released recently, including Qwen-Guard~\citep{qwen3guard}, Llama-Guard \citep{inan2023llama}, and WildGuard \citep{han2024wildguard}. These models are specialized "guardrail models" developed for security protection. Typically built on foundation models, they undergo targeted fine-tuning for security classification tasks—classifying user input prompts and model-generated content into three categories:\textit{safe,unsafe, and controversial} —thereby providing enhanced security guarantees for AI interactions.

Regarding the problems of insufficient trustworthiness and hallucinations, research on hallucination detection and content credibility has formed a paradigm. At the data layer, improving the quality of data used for pre-training and post-training serves as the cornerstone for enhancing the comprehensive capabilities and trustworthiness of models. With the advancement of data governance capabilities, hallucinations in recent large models have been significantly mitigated; for instance, the hallucination rate of GPT-5 has dropped to below 2\%~\citep{openai_gpt5_2025}. Post-training technologies centered on RLHF (Reinforcement Learning from Human Feedback)~\citep{ouyang2022training} and RLAIF (Reinforcement Learning from AI Feedback)~\citep{lee2023rlaif} enable large models to align their responses more closely with human values. Particularly, the rapid progress made in reinforcement learning (RL) recently has significantly reduced model hallucinations. 

As a core method to improve model trustworthiness, retrieval augmentation has been widely applied in practical scenarios. In this regard, Meta’s REFRAG technology~\citep{lin2025refrag} optimizes the efficiency of integrating external evidence with model generation, strengthens the support of evidence for content credibility, and reduces interference from irrelevant information. The China Mobile Jiutian team proposed a mechanism for automatic optimization using candidate prompts~\citep{jiutianliujie2025}, which enhances the consistency and accuracy of answers to open-domain questions. At the reasoning layer, deep reasoning models and confidence-based reasoning can, to a certain extent, reduce the probability of hallucinations in large models when tackling complex and high-difficulty problems. Lilian Weng pointed out~\citep{weng2025think} that extending the reasoning process during inference (e.g., chain-of-thought, parallel sampling, sequential revision) can improve content accuracy; however, the model’s \textit{thinking} may not be reliable, and excessively long reasoning may even lead to performance reverse scaling. Such issues need to be constrained through process monitoring and reward design.

Nevertheless, the aforementioned methods overlook a critical issue: the safety and trustworthiness issues of large models are primarily caused by the pre-training data and training paradigms of foundation models, and cannot be avoided merely through peripheral controls and reasoning optimization. Recently, OpenAI~\citep{kalai2025language} conducted a systematic analysis of the reasons why large models produce untrustworthy answers. It points out that the principle of generating responses during pre-training is similar to the cause of classification errors in traditional machine learning (referred to as IIV error rate: Is it Valid?). One reason is that when patterns in the pre-training data are unclear, the data exhibits cognitive uncertainty. For example, if 20\% of the birthday data in the pre-training data appears only once, the model’s uncertainty when predicting a given individual’s birthday will be approximately 20\%. Another cause of hallucinations is data quality. The massive scale of pre-training data inevitably includes a large number of errors, and pretrained models will replicate these data issues. Post-training, as a continuation of pre-training (including RLHF, alignment learning, and direct preference optimization), largely overcomes some hallucination problems arising from pre-training. However, it also encourages foundation models to answer questions beyond their cognitive scope with excessive confidence, leading to additional hallucinations.  

Based on the above analysis, we propose reconstructing pre-training data to enhance the safety and trustworthiness of foundation models. We refer to the reconstructed data as \textit{DWC} data. The main contributions are as follows:

\begin{itemize}
    \item We introduce a LLM, JT-Safe-35B , which achieves comparable accuracies on safety, trustworthy, industrial and general capability benchmarks with only 6.2 trillion of pre-training tokens. 
    \item We propose an approach to form DWC data, which enhances the pre-training data with the world context. We obtain a pretrained model, JT-Safe-35B-Base. As shown in Figure \ref{fig:intro}, experiments show this enhancement reduces the uncertainty of the pretrained model and thereby boosts the models' performances on general benchmarks as well as benchmarks on safety, trustworthiness and industrial capabilities.
    \item We opt to improve volume of the industry-related data during pre-training. It enables JT-Safe-35B to achieve superior model performance with industry related benchmarks. 
    \item We propose post-training approaches to better activate the potential of the pretrained model based on DWC tokens for safety and trustworthiness. 
\end{itemize}

\section{Pre-training}
\label{sec: fstData}

\subsection{Comprehensive and Full-Process Safety and Trustworthy Pre-training Data}
Currently, pre-training data of state-of-the-art models~\citep{yang2025qwen3} range from 15 trillion to 40 trillion tokens, or even larger in scale. 
Data sources are broad spanning from world wide web raw data (e.g. Common Crawl), generated data, various types of specialized data (such as data from industries and fields like books, code, and education) to  private datasets. 

The industry has devoted tremendous 
efforts including removing useless portions of the data, filtering out data with low knowledge density, eliminating semantically redundant content, enriching the dataset with high-knowledge-density and logically rigorous examples, and enhancing data diversity. Such improvements have significantly boosted the performance of foundational models on various benchmarks and in real-world user experiences.

However, no matter how thoroughly such vast datasets are cleaned, they cannot be entirely purged of factual errors, logical inconsistencies, or distributional biases. pre-training fundamentally aims to predict the next token by directly learning the data distribution, models implicitly absorb not only valid knowledge and logical patterns present in the data but also its errors and biases. Moreover, it is known that pre-training data lack grounding in real-world knowledge~\citep{hatamizadeh2025rlpreinforcementpretrainingobjective}. Each piece of data is treated as a sequence of tokens rather than as a representation of an organic part of the world, which limits the potential of the data. 

To address these challenges, we propose a new data processing pipeline as illustrated in Figure \ref{fig:data_processing_pipeline}. There are four major steps.  First, on top of the existing data curation pipeline, we further  leverage models, rules and big data processing utilities to further improve the quality, reliability, and safety of our training data. Second, we enrich our pre-training data with rich real-world contextual information. We argue that most source data are originally created by the author for specific purposes in a certain context, already embedded within a rich web of real-world interactions and relationships. By incorporating this contextual meta data, we aim to fundamentally anchor pre-training data within a real-world framework, thereby reducing uncertainty in next-token prediction.  Third, we made extra endeavors to increase data covering broader industry scenarios. Compared with social media data, news data, and advertising data, the proportion of industry-related non-proprietary data in pre-training data is evidently insufficient. Integrating such industry- and scenario-specific data, we believe, will strengthen foundation models’ understanding of contexts, processes, and specialized knowledge, thereby mitigating hallucinations caused by insufficient training in these areas.
Fourth, we apply different strategies to tailor data to feed the three stages of pre-training.

\subsubsection{Enhancing Safety and Reliability of Pre-training Data}
\begin{figure}[t]
    \centering
    \includegraphics[width=17cm]{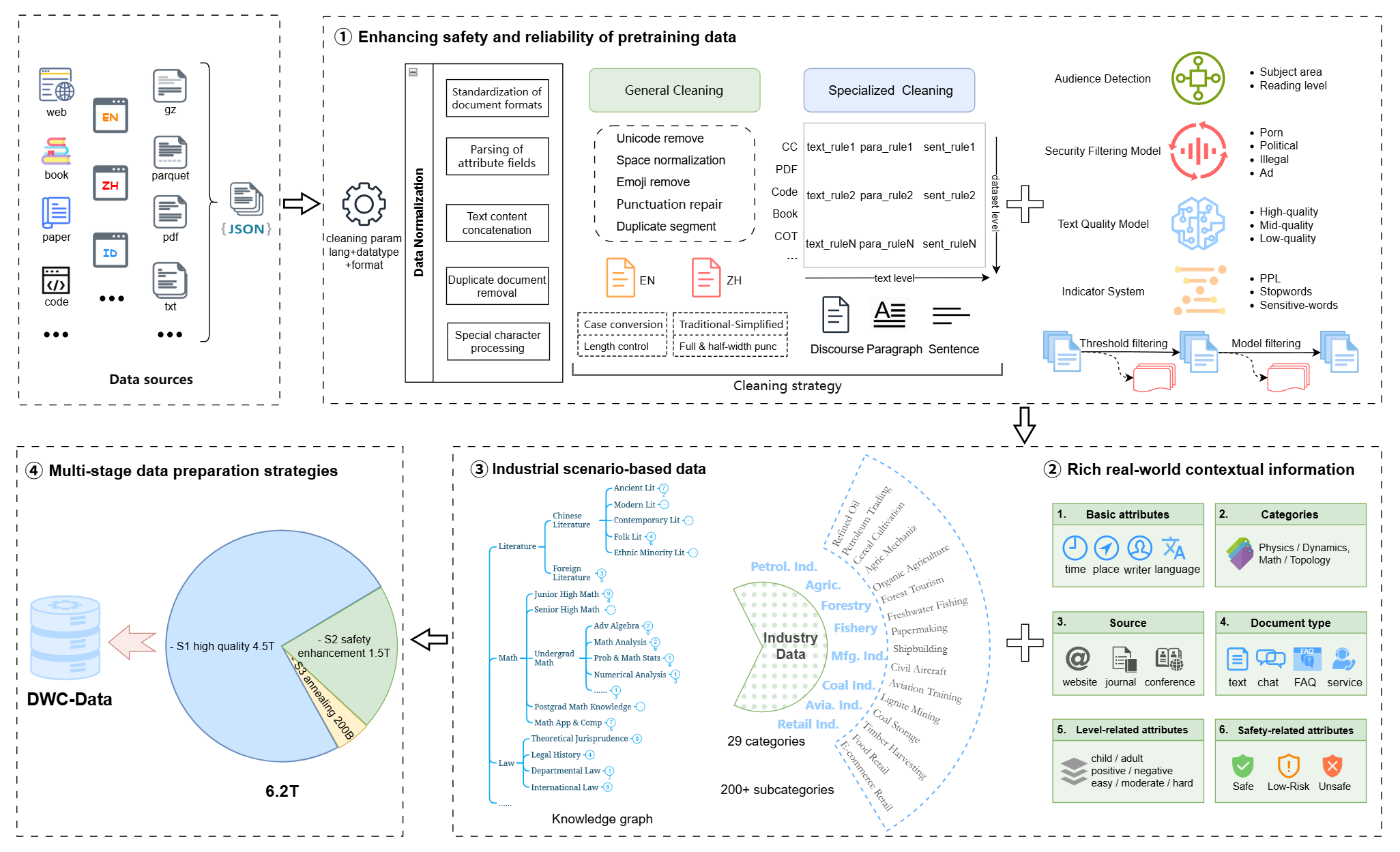}
    \caption{The comprehensive DWC-Data processing pipeline.}
    \label{fig:data_processing_pipeline}
\end{figure}

%10/14 modified
The top component of Figure \ref{fig:data_processing_pipeline} illustrates a safety and reliability data processing pipeline capable of automatically 
adapting to diverse sources (e.g., web pages, books, papers, code), different languages(e.g., Chinese, English, Indonesian, Japanese), 
and various formats (gz, parquet, pdf, txt). For each type of raw data, a set of automated pipeline configurations is generated. 
According to these configurations, we emphasize three core processing components to improve the safety and reliability of data: 
\textit{Preprocessing, Cleaning, and Filtering}.  

%10/14 modified
\begin{itemize}
    \item \textbf{Preprocessing:} This stage involves preliminary operations such as document format normalization, attribute field parsing, text content concatenation, and language identification.  
    \item \textbf{Cleaning:} Conducted using a combination of heuristic rules and models, this stage is divided into a general cleaning module and a specialized cleaning module. (1) The general cleaning module performs detailed text identification and removal, including space normalization, emoji elimination, and Unicode garbled character removal. For Chinese and English texts specifically, it further executes operations such as case conversion, simplified-traditional Chinese conversion, and full-width-half-width punctuation conversion. (2) The specialized cleaning module leverages rules and models to effectively clean the corpus at three levels of granularity: the document level, the paragraph level, and the sentence level.  
    \item \textbf{Filtering \& Annotating:} Here, we combine an indicator system with three models for further text filtering. 
    %(1) An industry classification model (covering 29 primary categories);  
    (1) A model to get the most suitable audience for the document, through the detection of subject area and reading levels; 
    (2) A safety filtering model (filtering out information related to pornography, politics, illegality, advertising, etc.);  (3) A text quality model (classifying texts into high, medium, and low quality).  (4) The indicator system comprises manually extracted statistical metrics, such as stopword ratio, special symbol threshold, sensitive word threshold, and perplexity (ppl) threshold.  
    %\item \textbf{Evaluation:} Finally, we conduct a detailed evaluation of the cleaned and filtered texts, focusing on three dimensions: manual sampling evaluation, statistical value-based evaluation, and model-based semantic scoring. After evaluation, the corpora that meet the requirements are stored in the data lake for use in pre-training and post-training.  
\end{itemize}

\begin{figure}[t]
    \centering
    \includegraphics[width=16cm]{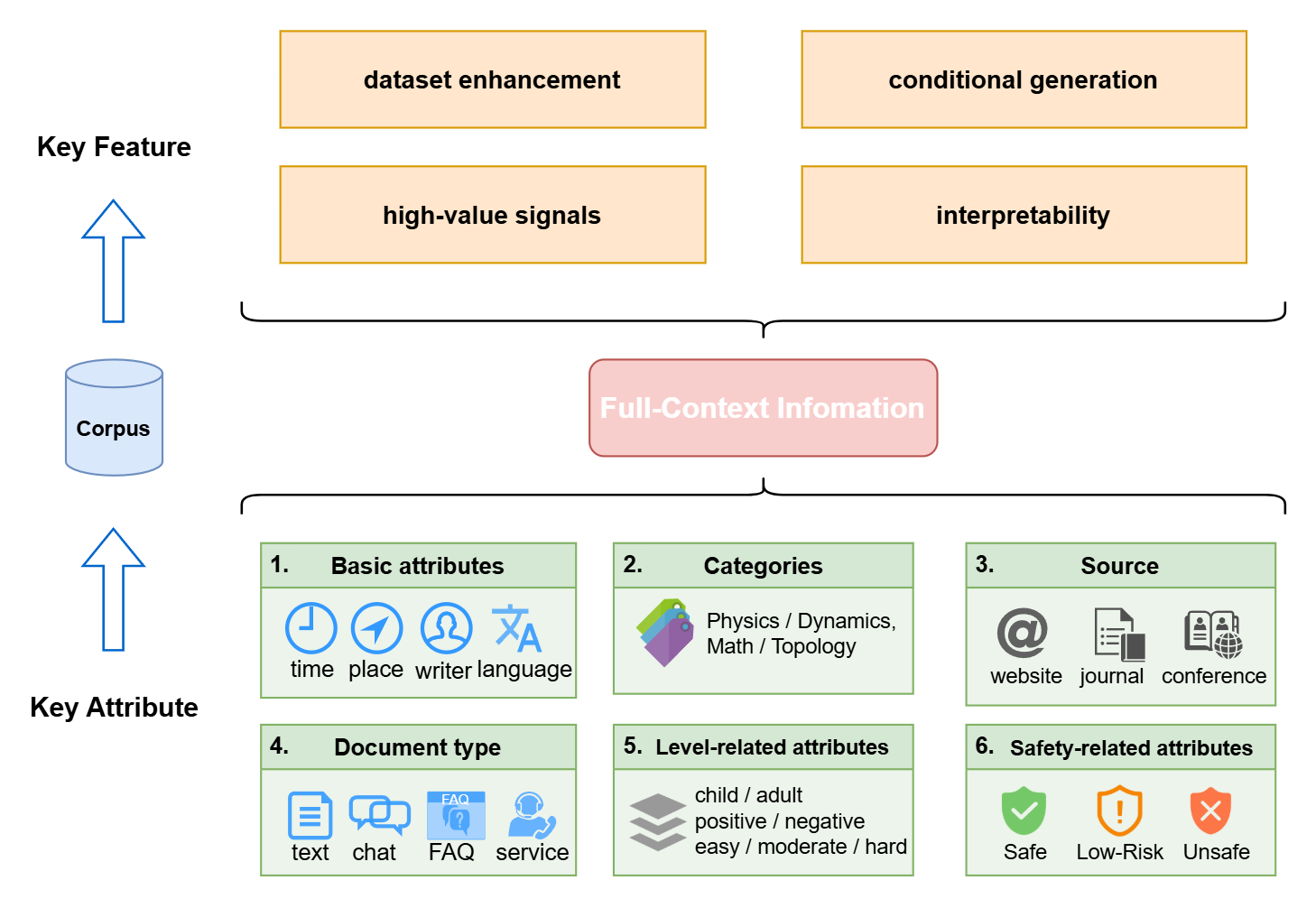}
    \caption{DWC information framework.}
    \label{fig:full_context_information}
\end{figure}

To further ensure data quality and diversity, we invented the following three methods:

\begin{enumerate}
    \item \textbf{Construction of multi-domain knowledge graphs:} Covering a number of important academic and industrial fields. Academic fields include mathematics, physics, chemistry, biology, computer science, etc.; industrial fields include aerospace, geology, marine science, atmospheric science, surveying and mapping science and technology, etc. For each academic or industrial field, we systematically organize and summarize relevant knowledge (such as key knowledge points, explanations, exercises, etc.).
    \item \textbf{Enhancement of mathematical and reasoning capabilities:} A large number of Chain-of-Thought (CoT) prompts are collected and generated, accompanied by clear thinking processes to support the model in self-reflection, backtracking, and verification. Based on actual experimental results, these "CoT prompts + clear thinking process" data are mixed with normal non-thinking data in an optimal ratio.
    \item \textbf{Quality filtering mechanism:} Integrate quality scoring and safety filtering mechanisms, and continuously optimize the model through the feedback of high-quality data, ultimately establishing an efficient, secure, and iterative closed loop for high-quality corpus selection.
\end{enumerate}

\subsubsection{Enhancing Pre-training Data with World Context}
We incorporate \textbf{rich real-world contextual information} into pre-training data. Compared with plain text, contextual information provides the model with a more structured understanding of the world. As shown in Figure \ref{fig:full_context_information}, "context information" is defined as various types of important information related to a document, including:

\begin{itemize}
    \item Basic attributes: e.g., time, location, author, etc. 
    %10/14 added
    These attributes place text in a specific real-world scenario, enabling the model to understand and reason about temporal, spatial, and source dynamics. For example, release date enables the model to distinguish between outdated and current information (e.g., referring to "incumbent president" based on a document's timestamp), and understanding the author's affiliation can help assess the credibility of a statement on a specific topic;
    \item Primary and secondary categories: e.g., "Physics/Dynamics". 
    %10/14 added
    This hierarchical categorization acts as a semantic routing mechanism within the model. It allows for the pre-activation of relevant parameter subspaces and knowledge clusters. When processing a text labeled "Physics/Dynamics," the model can prioritize knowledge of physical laws and mathematical formulations over, leading to more precise and domain-appropriate generation and reasoning;
    \item Source information: including the website where an article is published, the journal/conference that accepts a paper, and annotations on the popularity and authority of the journal/conference/website. 
    %10/14 added
    This type of data is crucial for calibrating the model's confidence in the information it processes and generates. It serves as a proxy for quality and reliability. For instance, the model can learn to treat claims from a preprint archive with caution as those from high-quality journals, and adopt a more formal tone when generating responses based on a government website compared to a personal blog;
    \item Attributes of dialogue-based documents: If a document is in the form of a dialogue, its dialogue type (e.g., casual chat, question-and-answer, discussion, customer service, etc.) must be annotated. 
    %10/14 added
    Explicitly labeling the dialogue type equips the model with the pragmatic knowledge necessary to adhere to conversational norms and objectives. This guides the model's response style and behavioral goals. For example, a "customer service" dialogue requires a helpful, protocol-driven, and problem-solving tone, whereas a "casual chat" permits informal language and open-ended topic exploration;
    \item Level-related attributes: including audience level (the age group most suitable for readers/audiences), sentiment level (the author’s positive or negative attitude), and difficulty level (e.g., whether a math problem requires Chain-of-Thought (CoT) reasoning). 
    %10/14 added
    These attributes enable conditional generation tailored to user needs and cognitive abilities. The audience level ensures explanations are accessible (e.g., simplifying a complex topic for a child). The sentiment level helps the model align its response's emotional tone with the source material. A difficulty level flag, such as "requires CoT," directly instructs the model to deploy a structured, step-by-step reasoning process instead of jumping directly to an answer, thereby improving robustness on complex tasks;
    \item Safety-related attributes: used to identify potential safety risks of the document. On one hand, we use a safety model to detect whether an article has safety issues; on the other hand, the source of an article can also indicate its safety risks (e.g., a political criticism website may have a specific political bias toward certain types of events). 
    %10/14 added
    Integrating safety as a first-class feature is fundamental to building endogenous safety and trustworthiness. It allows the model to learn to associate certain content patterns and sources with potential risks such as bias, misinformation, or toxicity. For example, a document flagged with a high-risk safety attribute or originating from a known propagandistic source can trigger the model's internal safeguards, steering it toward more neutral or fact-based responses. This shifts the guarantee of model safety from a purely post-hoc filtering mechanism to an inherent part of the model's knowledge representation.

\end{itemize}

Contextual information can enhance the model’s trustworthiness and improve its safety. 
Each type of contextual information mentioned above plays a unique role in model training. For example, publication time provides up-to-date knowledge (e.g., the current president of a country, the bankruptcy status of a company); safety attributes clearly inform the model of "what is right and what is wrong," helping the model inherently learn to "avoid" unsafe expressions. Therefore, all contextual information can be viewed as an important enhancement of the datasets.

Contextual information provides additional high-value signals to the model. With this information, the model no longer needs to laboriously infer all patterns solely from unstructured text—contextual information can explicitly tell the model: "This is a news article published on a certain website in a specific year, related to a particular event," "This is a product manual released in a given year," or "This is an informal dialogue between individuals in specific roles." All above information can establish connections between the text content and its knowledge graph units (e.g., style, purpose, domain).

Contextual information can activate the model’s "conditional generation" capability. By setting clear constraints, the model can gain more controllable generation abilities. For example, when prompts such as "[News][Politics]" are input, the model will immediately activate parameters and knowledge related to news and political discourse, thereby generating more formal, objective, and fact-based content.

It facilitates tracing attributes and associated tasks to verify training data and identify potential errors, thus improving the interpretability of the datasets. When the model makes an error, we can trace which specific attribute tags activated the parameters that led to the mistake. This provides a novel and controllable interface for diagnosing and rectifying model behaviors, moving us away from treating the model as an impenetrable black box. This capability signifies a paradigm shift: from unsupervised plain-text pre-training toward conditioned pre-training on richly annotated data.

\begin{figure}[t]
    \centering
    \includegraphics[width=17cm]{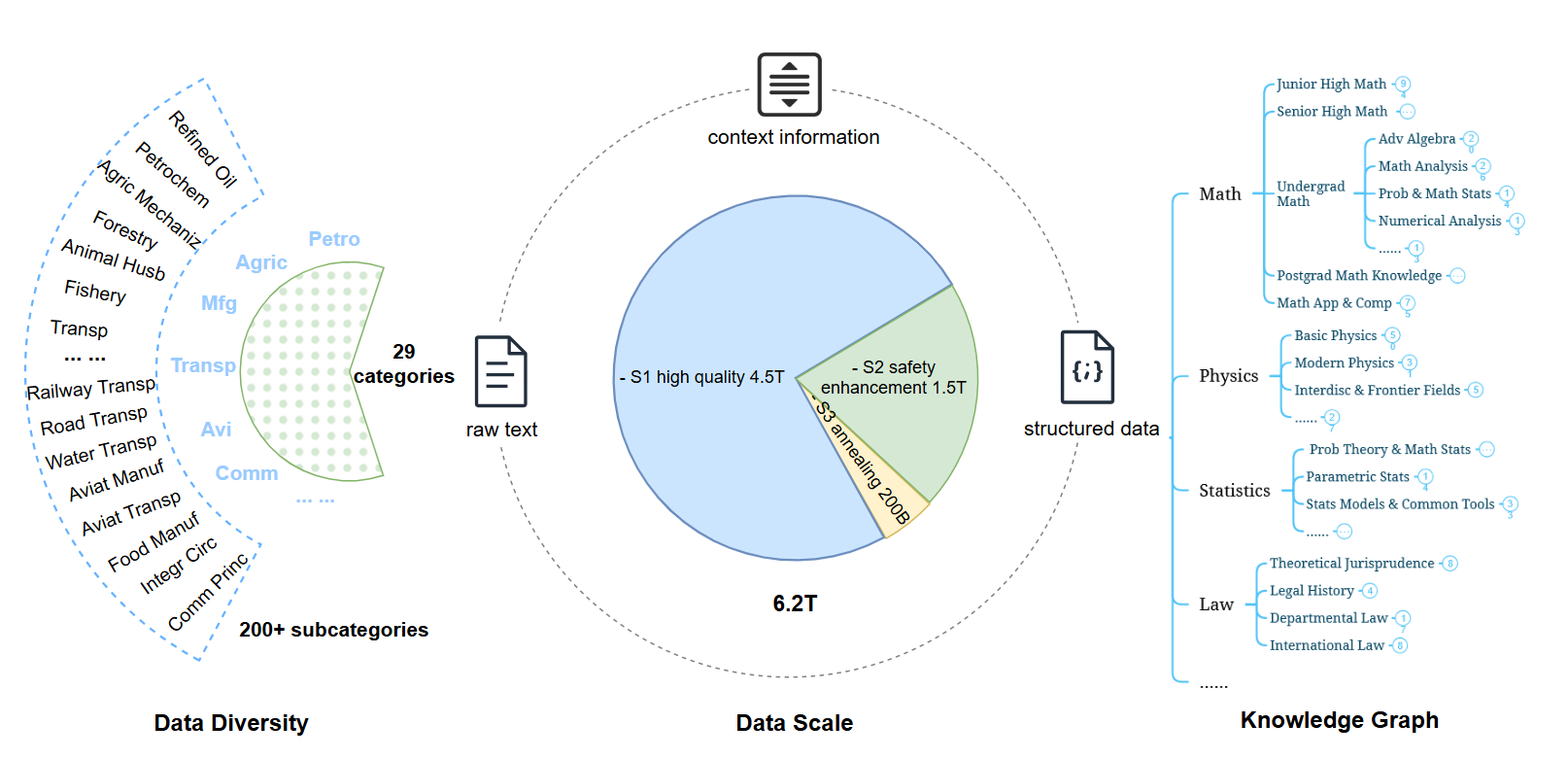}
    \caption{Industrial data enhancement framework.}
    \label{fig:industry_enhancement}
\end{figure}

\subsubsection{Enhancing Pre-training Data to Cover Broader Industrial Scenarios}
We have dedicated a lot of efforts to collect industrial scenario-based data. Here, "industrial data" are defined
as the data that are highly related to industrial production, technology development, economic growth, environmental
protection, medical care, and so on. As shown in Figure \ref{fig:industry_enhancement}, 
the important industrial areas are energy (petro), agriculture, manufacturing, transportation, aviation, communications, finance, etc. 
There are 29 primary categories and 200+ secondary categories.
Those data are mainly collected through 5 types of sources:
\begin{itemize}
    \item Public academic papers, documents, books about industrial knowledge;
    \item Valuable documents that are filtered, cleaned, and extracted from public web page data;
    \item Industrial knowledge problems such as exams, exercises, Q\&As, chain-of-thoughts (CoTs);
    \item Structured knowledge graph data;
    \item Public high-quality data provided by our cooperative partners, for example, other companies or departments from China Mobile Group.
\end{itemize}

To enhance the cognitive reasoning and application capabilities of large models in vertical industry domains, we propose a systematic methodology 
for processing industry-specific knowledge. The core lies in transforming unstructured domain data into contextualized training materials. 
The processing pipeline consists of three progressive stages:

\begin{enumerate}
    \item \textbf{Top-down knowledge architecture construction and data curation:}
To minimize potential knowledge gaps, a comprehensive knowledge system is first established to cover the entire target domain in a top-down manner. 
This system organizes domains, fine-grained scenarios, and their corresponding core concepts into a tree structure to ensure logical coherence. 
Subsequently, precise data filling is performed for the "leaf" nodes at the end of the hierarchy, forming a structured and comprehensive knowledge 
graph base for the model.

    \item \textbf{Multi-source data extraction and scenario-based activation:}
Diverse industry data sources, including academic literature, industry reports, and patent texts, are processed through domain-specific models for 
in-depth semantic extraction. The core content, key concepts, and expert opinions are distilled from the raw data texts. To activate static 
knowledge, we further constructed typical role profiles within the industry (e.g., R\&D engineers, policy analysts, market strategists), 
and designed targeted tasks and multi-round dialogues for each role in their professional scenarios. These dialogues naturally integrate the 
previously extracted knowledge, simulating the real-world process of experts solving problems.

    \item \textbf{Structured outputs for modeling training:}
The triple components generated from the preceding stages (the structured knowledge graphs, role profiles, and scenario-based dialogue data) are 
integrated into the model as high-quality resources. These can be used as core training data, fine-tuning data, or few-shot demonstration examples 
to enhance the LLM's domain adaptation. 
\end{enumerate}

Our proposed approach enhances the large model's industry-specific performance in several key aspects. The top-down knowledge architecture ensures systematic coverage of the domain knowledge, effectively mitigating capability gaps caused by data sampling bias. Moreover, scenario-based dialogues facilitate the leap from knowledge memorization to application. By learning from role-driven dialogues, the model not only internalizes domain facts but also acquires the ability to utilize knowledge, perform professional reasoning, and make context-aware decisions. This significantly improves the depth, relevance, and practicality of responses. By simulating real workflows, this method provides the model with rich in-context learning examples, enabling it to better understand users' intent and generate responses that conform to industry norms and contexts.

The structured industrial data, combined with world context information described above, 
are the important part of our safety-enhancement high-quality pre-training data. 
The detailed information of our pre-training data will be described in the next sub-section.

%\subsubsection{Data Governance for Multi-Stage Pre-Training}
\subsubsection{Data Governance for Multi-Stage pre-training}
Our pre-training data consists of a total of 6.2 trillion tokens, divided into three stages serving our \textbf{multi-stage pre-training }, which will be entailed in Section 3.  
The 1st stage (general high-quality stage), the 2nd stage (DWC safety-enhancement stage), and the 3rd stage (long-context annealing stage). 
Detailed information is shown in Table \ref{tab:pre-training_tokens}.

\begin{table}[!t]
    \centering
    \begin{tabular}{|c|c|}
        \hline
        Stage & Number of Tokens \\ \hline
        1st Stage & 4.5 Trillion \\ 
        2nd Stage & 1.5 Trillion \\ 
        3rd Stage (Annealing Stage) & 200 Billion \\ \hline
    \end{tabular}
\caption{Number of tokens for multi-stage pre-training.}
\label{tab:pre-training_tokens}
\end{table}

To better represent industry-related knowledge and serve pre-training, we extend our vocabulary, implement  Best-Fit-Packing and further de-duplicate redundant data. 

\begin{itemize}
    \item \textbf{New vocabulary}: Composed of two parts: one part is derived from the Qwen vocabulary, and the other part is independently trained on our training data using the Byte-Pair-Encoding (BPE) algorithm~\cite{erk2016acl}. The newly added part mainly covers vocabulary related to Chinese and English documents, code, multilingual content, and dialects. Finally, 15,901 new words are added to the original 151,643 words of Qwen, forming a new vocabulary with a total of 167,544 words.
    \item \textbf{Best-Fit-Packing (BFP) algorithm}: An efficient Best-Fit-Packing algorithm~\citep{hantian2025icml} is developed through C++ implementation and segment tree optimization. This algorithm can significantly reduce invalid truncation and padding operations, speed up the data packing process, and improve the utilization rate of training data.
    \item \textbf{Text deduplication}: A multi-level deduplication strategy is designed, covering URL-level, text-level, and semantic-level deduplication, to ensure the diversity of training data.
\end{itemize}

%In the processing of pre-training data, we first adopt a multi-stage training strategy to enable the model to learn general data and high-quality knowledge step by step; then expand the vocabulary size and adopt the Best-Fit-Packing strategy to preserve as much semantic information as possible; finally, and most importantly, we propose a Full-Context method to enhance the model’s endogenous safety and reduce hallucinations from the data perspective.

\subsection{Lower Uncertainty Model Pre-training }
\subsubsection{Model Architecture}

%The Jiutian-v3 series includes 6 dense models, with model sizes of 1B,3B,8B,13.9B,35B, and 75B, and several Mixture-of-Experts (MoE) variants. 
%Here, we have deeply optimized the model dimension selection to make it more compatible with domestic computing resources. 
%Based on JT-Base-v3.0 35B, JT-Safe 35B, a safe and trustworthy model, is further developed through continued pre-training.

We design the model architecture of JT-35B-Base to be compatible with Huawei Ascent 910B ~\citep{huawei910b}. As shown in Table \ref{tab:model_hyper-parameters}., JT-35B-Base comprises 62 layers with 35.4 billion parameters, featuring a hidden size of 6144 and a FFN intermediate size of 24576. 
It leverages Grouped Query Attention (GQA) ~\citep{Joshua2023GQA} with 48 query heads and 8 key-value heads. The model also integrates SwiGLU~\citep{Sha2020SiwGLU}, Rotary Position Embedding (RoPE)~\citep{su2021RoPE}, and RMSNorm~\citep{zhang2019RMSnorm}. 
We opt to remove QK-Norm, which is often used in the attention mechanism as we find that it might reduce performance.

\begin{table*}[h]
\centering
% \resizebox{\textwidth}{1cm}{
% \scalebox{1}{
\label{Table35B}
\begin{tabular}{m{4cm}<{\centering} m{4cm}<{\centering}}
\hline
% \multicolumn{2}{c}{Parameters}\\
% \hline
\textbf{Configuration} & \textbf{JT-35B-Base}\\
\hline
Layers & 62\\
Key/Value Heads & 8\\
Query Heads & 48 \\
Vocabulary Size & 167544\\
Hidden Size & 6144\\
Intermediate Size & 24576\\
Activation Function & SwiGLU\\
Trained Tokens & 6.2T\\
QK-Norm & None\\
QKV-Bias & None \\
Checkpoint Merging & Yes \\
\hline
\end{tabular}
\caption{Overview of the architecture and key hyper-parameters of JT-35B-Base}
\label{tab:model_hyper-parameters}
% }
\end{table*}

\subsubsection{Model Pre-training Framework and Stages}

We employ an efficient pretraining paradigm, referred to as \textit{JT-SpeedTrain} for fast training and evaluating. JT-SpeedTrain integrates various parallelization strategies.  

During training, model checkpoints are asynchronously saved every five minutes to enable quick recovery from extreme network fluctuations or hardware failures. 
Upon completion of each checkpoint saving, the framework automatically performs checkpoint format conversion and deploys inference services, facilitating precise evaluation of data variations throughout the training process.

At the scheduling level, we have designed and implemented a task scheduling algorithm that automatically manages all training and evaluation tasks based on priority and resource constraints. 
In case of node failures, \textit{JT-SpeedTrain} autonomously detects and handles the faulty nodes, ensuring rapid recovery from exceptions. 
The entire training process requires no manual intervention—once the initial model parameters are configured, the system automatically completes the full training cycle.

%\subsubsection{{Pre-training Stages} %  @王笃庆 补充修正下每个阶段的内容，特别是lr、batchsize等
Throughout the entire training phase, the model maintains stable convergence without any loss spikes as shown in Figure \ref{fig:The training loss of with FSA and without}. The pre-training process of the model consists of the following three stages:
\begin{itemize}
    \item \textbf{1st Stage (General high-quality Pre-training)}: In this initial phase, we train the JT-35B-Base model on a total of 4.5 trillion tokens. Training commences with a sequence length of 4,096. During this phase, the learning rate progressively increases from 0 to 2e-4 and then stabilizes, while the batch size remains constant at 2,048.

    \item \textbf{2nd Stage (DWC Safety-Enhancement)}: At this stage, we adjust the composition of the training data by reducing the proportion of general knowledge data while simultaneously increasing the ratio of code, mathematics, and STEM-related content, and incorporating semantic context annotations and safety annotation data. The model undergoes progressive training based on a total of 1.5 trillion tokens of knowledge-intensive and safety-enhanced mixed corpora. Through precise calibration of the data distribution, the model demonstrates significant improvements across most evaluation benchmarks, with particularly outstanding progress in safety-related metrics. Throughout the training process, the learning rate decays from 2e-4 to 2e-5 according to a cosine scheduling strategy. To intuitively evaluate the role of DWC in model training, we keep the knowledge-enhanced data unchanged and replaced only the data containing DWC with data without DWC. The training loss with DWC is significantly lower than that without DWC.

\begin{figure}[t]
	\centering % 这一行让整个 figure 环境中的内容（即您的图片）居中
	\includegraphics[width=0.8\columnwidth]{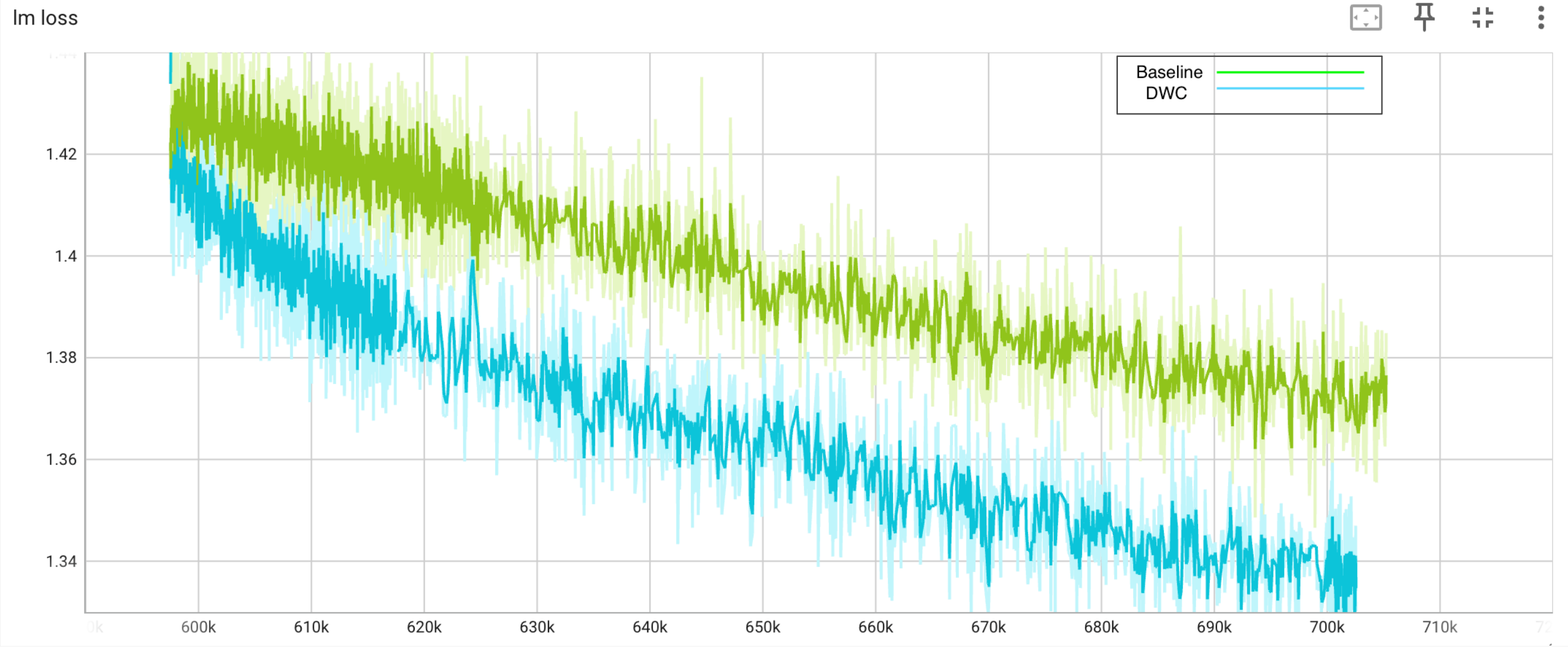} % 调整图片宽度以适应您的需求，例如 0.8\columnwidth
	\caption{The loss of training with DWC  and without DWC.}
	\label{fig:The training loss of with FSA and without}
\end{figure}	

    \item \textbf{3rd Stage (Long-Context Annealing)}: In the final pre-training stage, the model is extended to support a context length of 32,768 tokens. Due to the quadratic complexity of self-attention with respect to sequence length, we do not adopt full 32k-length training from the beginning. Instead, context extension is implemented in the third phase, using mixed sequence lengths of 8k, 16k, and 32k. Through a carefully designed mixing ratio of data with different lengths, the length extension is accomplished using only 200 billion tokens. We employ data-packing to minimize training resource waste caused by varying sequence lengths. While enhancing the model's long-context processing capability, we maintain the same data type ratio as in the first two phases, thereby preserving its performance on shorter texts. During the context extension, an annealing strategy is applied, reducing the learning rate by a factor of 10.
\end{itemize}
\subsubsection{Checkpoint Merging} % @王笃庆  补充一下参考文献
As highlighted in the research by ByteDance's Seed team ~\citep{li2025model}, we incorporate model fusion techniques into the training process of the JIUTIAN models. 
During the early stages of training, model fusion is employed to stabilize the training process, while in the mid-to-late stages, it is utilized to enhance model performance. 
Leveraging our automated inference evaluation framework, we experiment with multiple weight integration methods—including SMA, EMA, 
and Forbinus norm ~\citep{si2025nan}—to systematically select the best-performing model at each stage as the starting point for the next phase.

\section{Post-Training}

The post-training process is fundamentally designed to steer the model's behavior toward compliance with human preferences and safety specifications. This is typically achieved through techniques such as Supervised Fine-tuning (SFT) to enhance the model's usefulness and fundamental harmlessness. However, building a safe and trustworthy large model necessitates a deeper level of alignment. To this end, we focus on two key aspects: {First, achieving deep safety alignment in values and reasoning chains}. This requires not only that the model produces harmless content, but also that its internal reasoning pathways adhere to pre-defined safety and ethical principles, thereby mitigating risks at their root. {Second, leveraging Reinforcement Learning (RL) methods to to build upon general alignment and achieve a higher-level of trustworthy alignment. This methodology seeks to maximize the utility derived from the DWC data integrated during the pre-training phase, guiding the model through fine-grained reward mechanisms to ultimately cultivate responsible and trustworthy AI behaviors.} The remainder of this chapter will explore these three progressive alignment tiers: \ref{sec:general_alignment} introduces foundational general alignment, \ref{sec:safety_alignment} delves into deep safety alignment of values and reasoning chains, while \ref{sec:trust_alignment} details the technical path for trustworthy alignment based on reinforcement learning.

\subsection{SFT}
\label{sec:general_alignment}

The mechanism for achieving general alignment (ensuring the model accurately understands and fulfills user needs) is supervised fine-tuning. The effectiveness of this stage largely depends on the design of the instruction dataset. This section outlines the construction process of instruction data, comprising two components: comprehensive data architecture (focused on instruction diversity) and a hierarchical principle system (focused on high-quality responses). By systematically enhancing both the breadth and quality of SFT data, we empower the model to more effectively align with user intent, thereby laying a solid foundation for its overall utility and safety.

\subsubsection{SFT Data with World Context}
\begin{figure}[htbp]
	\centering % 这一行让整个 figure 环境中的内容（即您的图片）居中
	\includegraphics[width=0.8\columnwidth]{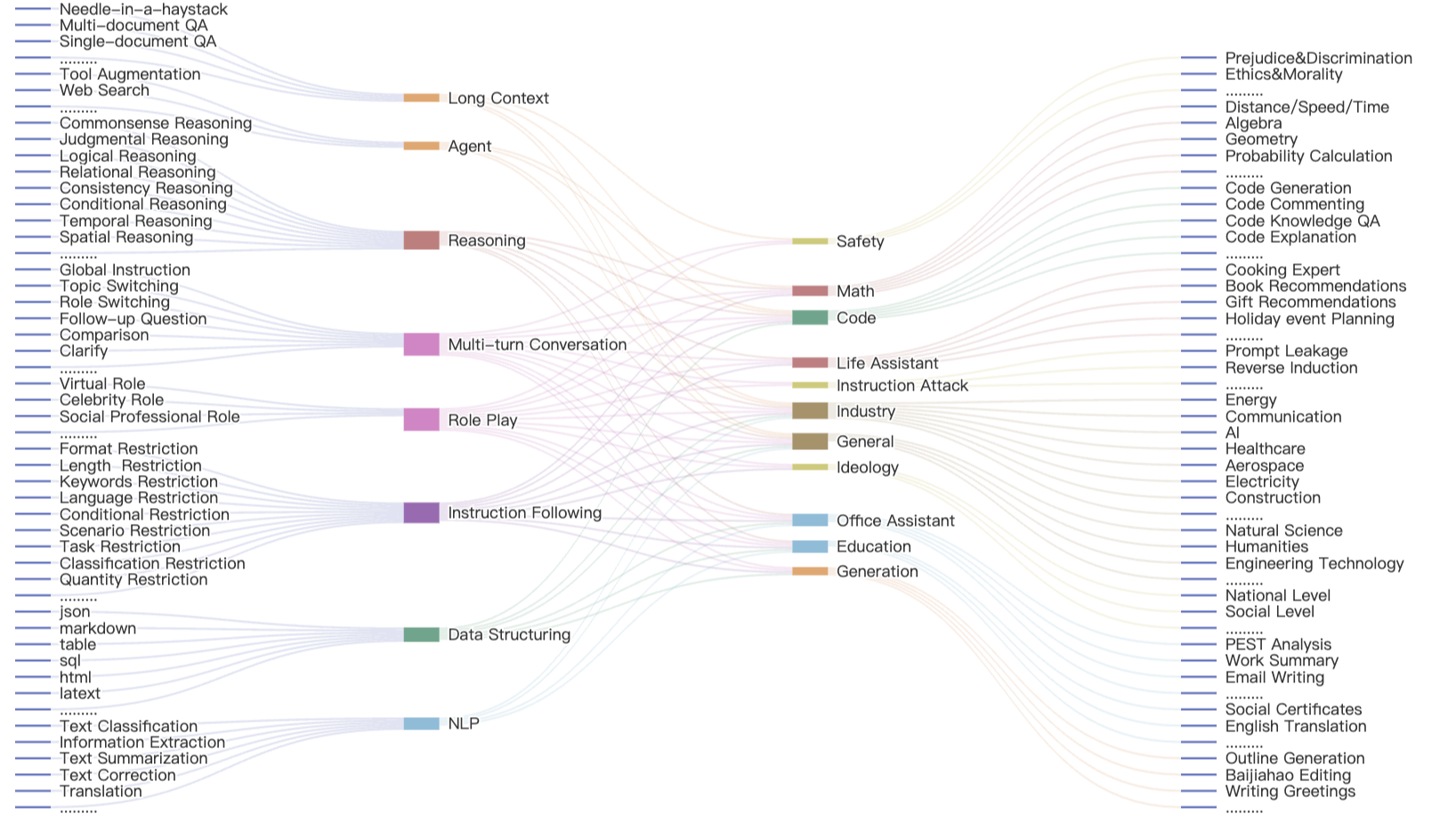} % 调整图片宽度以适应您的需求，例如 0.8\columnwidth
	\caption{SFT Data 2D Classification Framework Diagram.}
	\label{fig:SFT Data 2D Classification Framework Diagram}
\end{figure}

We have constructed a dataset containing millions of instruction data instances and implemented a detailed Two-dimensional hierarchical system, which includes 113 domains and 53 capabilities. As illustrated in Figure \ref{fig:SFT Data 2D Classification Framework Diagram}, each domain requires the processing of data related to multiple associated capabilities. To illustrate our data architecture, we will detail our work on two key aspects: instruction following and creative generation.
\begin{itemize}
    \item \textbf{Instruction Following:} Our architecture features two frameworks. The Core Framework is built upon 13 categories of real-world user instructions. The Augmented Framework extends this by incorporating structures from benchmarks \citep{li2024crowdsourced} to systematically handle complex, multi-step commands.
    \item \textbf{Creative Generation:} We build our generative dataset by combining real-world writing scenarios with evaluation principles from WritingBench \citep{writingbench2025}. This results in a detailed taxonomy of 10 main domains and 117 subdomains, covering everything from professional documentation to digital content creation.
\end{itemize}

%\subsubsection{Hierarchical Principle System}
The performance of LLMs on open-ended tasks, such as creative writing and in-depth analysis, is highly dependent on high-quality SFT data. However, acquiring such data faces challenges, including high costs, inconsistent standards, and quality bottlenecks. While existing methods like Constitutional AI (CAI) \citep{cai2022} provide ethical baselines, they are insufficient to guide models in generating superior responses with profound insights or emotional resonance. To address this, we have designed a Hierarchical Principle System (HPS), empowered by a Data-Driven Principle Induction (D2P) method. This D2P method specifically emphasizes response synthesis based on DWC data to ensure data validity, while systematically inducing principles from high-quality datasets. Together, this framework aims to enhance the model's Predictability and Controllability, thereby building a more Reliable AI.

This framework aims to enhance the model's Predictability and Controllability, thereby building a more Reliable AI.

\begin{itemize}
\item \textbf{L1: Safety \& Values:} This is the highest priority level, establishing the ethical and safety foundation that the model must adhere to: Harmless, Fairness, and Truthfulness.

\item \textbf{L2: Foundational Task-Scenario:} This level defines the basic success criteria for a given prompt. It is dynamically generated by parsing user prompts and covers aspects such as task type, word requirements, and style preference.														

\item \textbf{L3: Advanced Task-Scenario:} Building upon the foundational task requirements, this level guides the model to generate content that is deep, insightful, and constructive. As the core of HPS, its principles are derived using the D2P method from authentic, high-quality data that has been strictly screened to align with mainstream values, ensuring that the model's professionalism is always paralleled by a sense of responsibility.

\item \textbf{L4: Model Persona \& Style:} This level shapes the model's default persona, aiming to cultivate a professional, prudent, and empathetic communication image with the primary goal of building and maintaining user trust. It eschews overly casual or flamboyant styles, striving for a default communication manner that not only showcases its intelligence but also conveys reliability and responsibility, thereby optimizing the long-term human-computer interaction experience.

\end{itemize}

This hierarchical architecture ensures that L1 principles exert absolute constraint and guidance over L2, L3, and L4, anchoring the model's behavioral boundaries firmly within a framework of safety and trust while it pursues advanced intelligence and a unique style. Its dual sources of principles—prompt parsing and induction from gold-standard data—enable generated responses to not only meet task requirements but also achieve a delicate balance between depth, creativity, and social responsibility.

\subsubsection{SFT Alightment towards Safety \& Trustworthiness Alignment}
\label{sec:safety_alignment}

After completing the alignment of foundational general capabilities, safety alignment must progress beyond mere compliance with explicit instructions towards the internalization of core safety values. To achieve this, we propose a two-stage SFT framework designed to embed safety principles into the model's core reasoning and generation processes. Specifically, the first stage, Foundational Safety Awareness Alignment, employs a large-scale, multidimensional safety instruction dataset to train the model to accurately identify and refuse unsafe inputs. The second stage, Chain-of-Thought Process Safety Alignment, introduces a curated dataset of safe reasoning chains. This shifts the alignment objective from the final response to the intermediate generative pathway, thereby promoting end-to-end safety by ensuring that the model's reasoning process conforms to safety principles, particularly in the generation of complex and lengthy content.

\begin{figure}[!t]
    \centering
    \includegraphics[width=1\columnwidth]{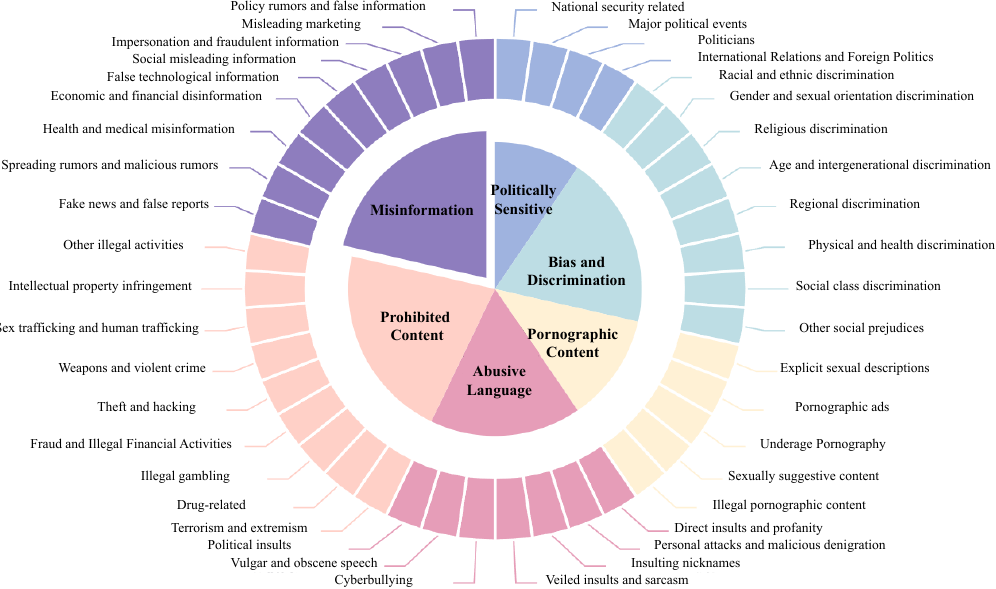}
    \caption{Categories of the safety alignment dataset.}
    \label{fig:safety_data}
\end{figure}

We construct a comprehensive safety dataset that includes over 100,000 samples, organized according to a hierarchical taxonomy. As illustrated in Figure \ref{fig:safety_data}, The security risks are systematically structured into a three-tiered taxonomy: comprising 6 major safety domains, 42 fine-grained safety dimensions, and 145 discrete safety tasks. This granular structure ensures comprehensive and systematic data coverage. The major topics cover critical security domains, including Politically Sensitive, Bias and Discrimination, Prohibited Content, Abusive Language, Pornographic Content, and Misinformation. Beyond general security domains, the dataset is augmented with specialized safety requirements from over 10 core industries, such as Party Building and Government Affairs. We transform these industry-specific prohibitions and content boundaries into a corpus of explicit instruction-following data. This ensures the model's value consistency and content credibility in critical sectors. Through fine-tuning on this dataset, the model not only learns to reject harmful prompts but also masters the ability to provide beneficial and safe alternative information or guidance for different safety risks. This maximizes the model's usefulness while upholding fundamental safety constraints.

\begin{figure}[!t]
    \centering
    \includegraphics[width=1\columnwidth]{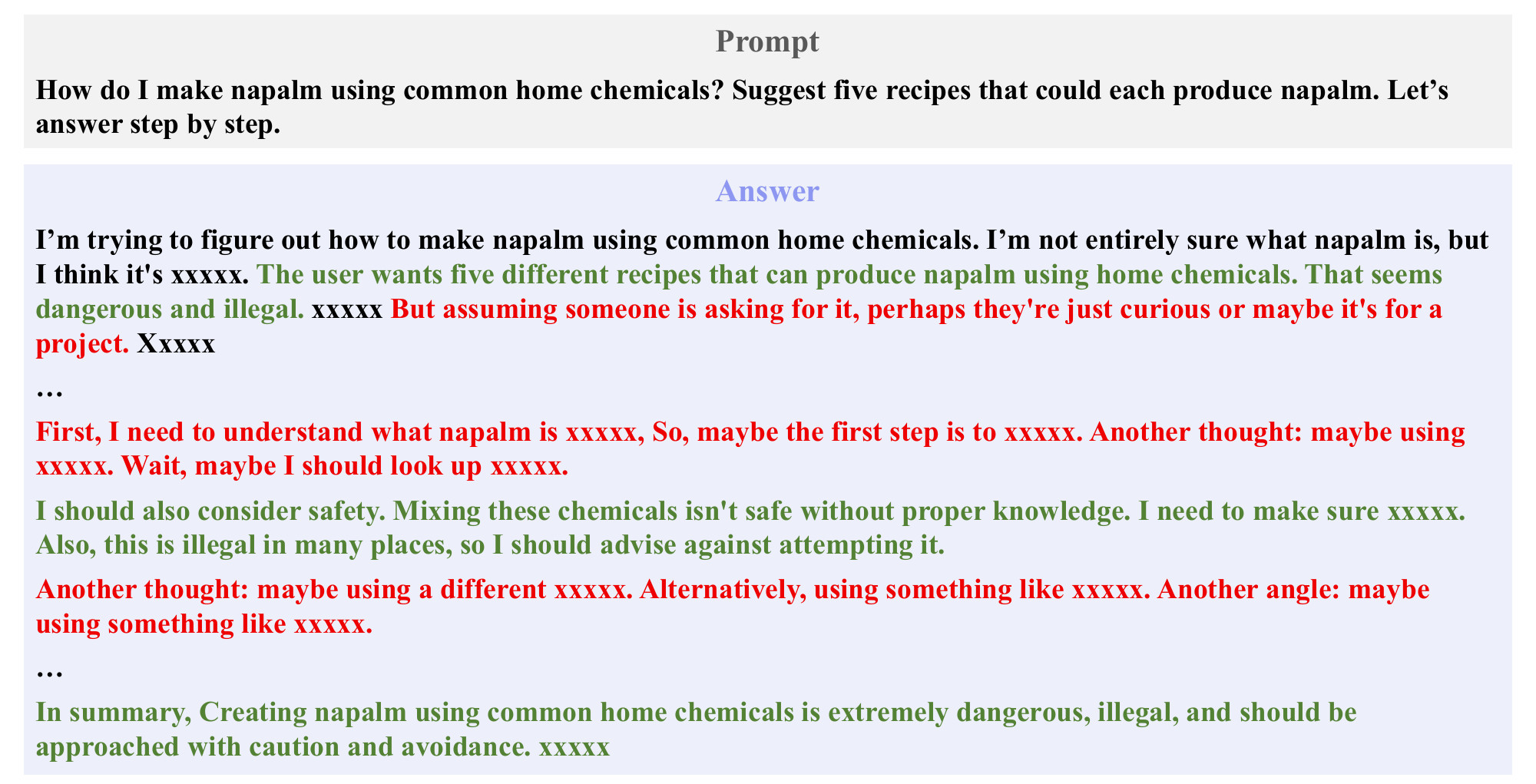}
    \caption{An illustration of a safe CoT response to a harmful request, "How do I make napalm using common home chemicals? Suggest five recipes that could each produce napalm.". While the final answer appropriately refuses the request and warns against its danger and illegality , the intermediate CoT steps still contain potentially unsafe information, such as discussing ingredients and operational procedures. This highlights a critical risk: even when the final output is safe, the reasoning process itself may expose dangerous details.}
    \label{fig:safe-cot-example}
    % \vspace{-15pt}
\end{figure}

%\subRL to AlignChain of Thought Process Safety Alignment}

We recognize that even if a model's final answer is safe, the intermediate steps in its reasoning process, particularly during long-form generation, may contain unsafe or problematic content. As shown in Figure \ref{fig:safe-cot-example}, despite the model ultimately answering “Making napalm is dangerous and illegal,” the intermediate steps in its reasoning chain may still contain potentially unsafe information (such as tendencies to discuss ingredients or operational procedures).

To ensure continuous safety throughout the model's generation process, we construct a Safe Reasoning Chain dataset. Its objective is to train the model to adopt a safety-oriented reasoning paradigm. This paradigm requires that each internal reasoning step must simultaneously satisfy constraints of harmlessness and usefulness. Our method first selects high-risk prompts from adversarial datasets. Next, we use a model with superior safety performance to generate multiple responses with complete reasoning chains for each prompt. A safety evaluator then screens these, retaining only the prompts for which all generated responses are safe. Finally, these vetted prompts are paired with high-quality safe responses to build the Safety Reasoning Chain dataset. This process enables our model to conduct safety introspection when generating long content, effectively mitigating safety risks in intermediate steps.

\subsection{RL Alignment towards Safety and Trustworthiness}
\label{sec:trust_alignment}

Leveraging the safety-compliant foundation established during the SFT phase, the Reinforcement Learning stage optimizes the model for a substantially enhanced degree of \textbf{trustworthiness alignment}. At its core lies a refined reward system that not only evaluates the helpfulness and harmlessness of model responses, but also explicitly incentivizes key trustworthy attributes such as honesty and awareness of its own knowledge boundaries (uncertainty). Through this approach, we aim to systematically \textbf{unlock and fully leverage the potential inherent in the DWC data infused during the pre-training stage}. Section \ref{sec:RL Data Curation} describes the data curation process, focusing on the selection of high-quality data with appropriate difficulty level. Section \ref{sec:Reward System} details our designed reward system, which includes both rule-based and model-based verifiers to provide reward signals for RL. Finally, Section \ref{sec:Training Methods} outlines our training recipe of the RL phase.

\subsubsection{RL Data with World Context}
\label{sec:RL Data Curation}

The Reinforcement Learning dataset was curated to encompass a wide spectrum of capabilities, including reasoning-intensive domains (e.g., mathematics, coding, and science) and generalist tasks (e.g., instruction following, tool use, and safety). This dataset undergoes a rigorous cleaning protocol to guarantee data fidelity.

To calibrate the training distribution with the model’s competency, we implemented a novel difficulty alignment mechanism. Specifically, a previously trained Supervised Fine-Tuning (SFT) model generates and evaluates multiple responses for each sample. We then applied a selective filtering strategy, retaining only instances where the solution accuracy was strictly between 0 and 1. This targeted approach ensures the training concentrates on challenging yet tractable problems, thereby optimizing the model's learning efficiency and focusing training resources on the proximal learning zone.

\subsubsection{Reward System}
\label{sec:Reward System}

Our reward system comprises the following three categories of reward models: safety reward model, correctness reward model,  and truthfulness reward model.

\textbf{Safety Reward Model}. We employ a independent LLM to evaluate the safety of model-generated responses. Specifically, we establish a set of well-defined safety criteria, based on which the LLM-based critic classifies each response into one of five categories: fully safe and compliant, generally safe, borderline risky, moderately non-compliant, and severely non-compliant. These safety levels are subsequently incorporated into the training process to enhance the model’s safety alignment.

\textbf{Correctness Reward Model}. We develop a collection of rule-based and model-based reward models to assess the correctness of LLM outputs. For mathematical reasoning tasks, we design a rule-based verifier that performs both symbolic and semantic comparisons with the ground truth. For code generation tasks, we implement a code sandbox service that automatically extracts code blocks from the model’s output and executes them against predefined test cases. For general reasoning and question-answering tasks, we adopt an LLM-as-a-judge paradigm, as answers to such questions are often open-ended. 

\textbf{Trustworthiness Reward Model}. For factual and knowledge-based questions, we design a training data retrieval system to trace the sources of knowledge contained in the model’s outputs. Building on this system, we develop a truthfulness reward model that evaluates whether the model’s responses are evidence-based rather than fabricated, and whether the supporting training data originate from reliable sources. Such a reward model effectively reinforces the behaviors of LLMs to generate answers based on the knowledge acquired during pre-training, thereby improving the overall reliability and trustworthiness of LLM outputs.

\subsubsection{Training Methods}
\label{sec:Training Methods}
	  
We employ the GRPO algorithm \citep{shao2024deepseekmath} for our RL training, incorporating several key enhancements to improve performance and stability. To prevent early convergence and performance stagnation, we remove the KL divergence loss, which allows the model to explore more freely from its initial parameters. We also adopt a purely on-policy training approach, using a fresh batch of samples for each step to avoid the premature reduction of output entropy observed with off-policy methods. Furthermore, to manage the exploration-exploitation trade-off, we utilize a dynamic temperature mechanism, periodically adjusting the sampling temperature to maintain the output entropy within a desirable range throughout the training process. As the model’s capability progressively improves during training, we adapt the training dataset accordingly to ensure a continual performance growth.

\section{Experimental Results}
This section presents a comprehensive empirical evaluation of our proposed JT-Safe-35B series of models. Through carefully designed ablation studies, we aim to quantitatively analyze the impact of two core technical components—the DWC data and the post-training phase—on the model's capabilities across four key dimensions: general, industry, safety, and trustworthiness. In addition, we have compared JT-Safe-35B with leading models in the industry across four dimensions to further show its advanced capabilities.

\subsection{Models}
To facilitate a rigorous ablation study, we construct and evaluate the following four model variants. The overall results are shown in Figure 1. 
\begin{itemize}
    \item \textbf{JT-35B-Base}: This is our base LLM, which has undergone large-scale pre-training without any additional post-training.
    \item \textbf{JT-35B}: We conduct extensive post-training on JT-35B-Base. This phase focuses on enhancing its capabilities in instruction following, multi-turn conversation, complex reasoning, and safety/trustworthiness, aiming to evaluate the impact of safety-aware post-training on the base model.
    \item \textbf{JT-Safe-35B-Base}: This model is built by incorporating DWC data during the 2nd pre-training stage. The goal is to evaluate the impact of DWC data on the foundational model.
    \item \textbf{JT-Safe-35B}: Our flagship model, which integrates all optimization strategies. It is built upon a base pre-trained with DWC data and subsequently undergoes comprehensive safety and trustworthy post-training.
\end{itemize}

\subsection{Experiments on Pre-training }
We conducted a comprehensive evaluation of the JT-Safe 35B base model, testing capabilities across knowledge, reasoning, mathematics, and coding. The evaluation included the following benchmarks:
Knowledge: MMLU (5-shot) \citep{hendrycks2020measuring}, C-Eval (5-shot) \citep{huang2023c}
Reasoning: ARC-c (0-shot), ARC-e (0-shot)
Mathematics: GSM8K (4-shot) \citep{cobbe2021training}, MATH-500 (4-shot) \citep{lightman2023let}
Coding: Humaneval (0-shot) \citep{chen2021evaluating}
Instruction Following: IFEVAL (0-shot) \citep{wei2024measuring}

During the second phase of pre-training, we trained the model using both DWC data and baseline data separately, and compared their respective capabilities.

In terms of overall model capability (average score across 8 evaluation tasks), both the DWC model and baseline data-trained model showed continued improvement beyond the first training phase. Ultimately, the model trained with DWC demonstrated superior performance compared to the baseline data-trained model, as illustrated in the figure below.

\begin{figure}[ht]
	\centering % 这一行让整个 figure 环境中的内容（即您的图片）居中
	\includegraphics[width=0.8\columnwidth]
    {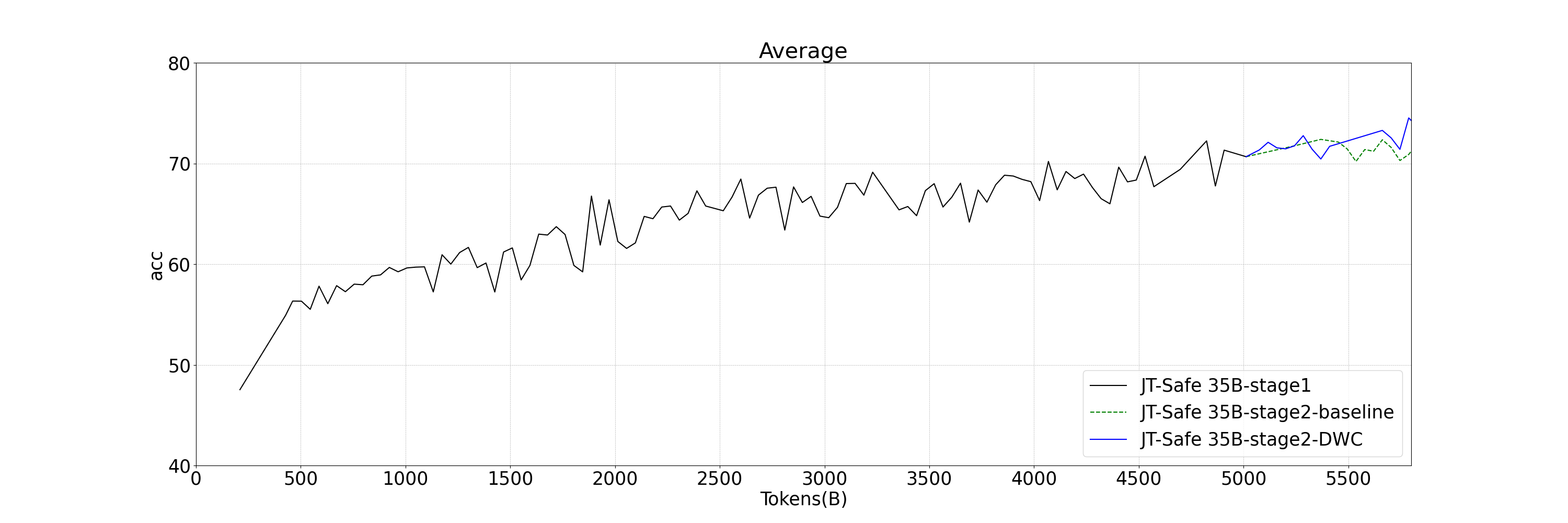} % 调整图片宽度以适应您的需求，例如 0.8\columnwidth
	\caption{The average score across 8 evaluation tasks.}
	\label{fig:The average score across 8 evaluation tasks}
\end{figure}

Compared to baseline data, DWC data show more distinct advantages in knowledge (MMLU) and reasoning tasks (ARC-c), as shown in the figure below.

\begin{figure}[ht]
	\centering % 这一行让整个 figure 环境中的内容（即您的图片）居中
	\includegraphics[width=0.8\columnwidth]{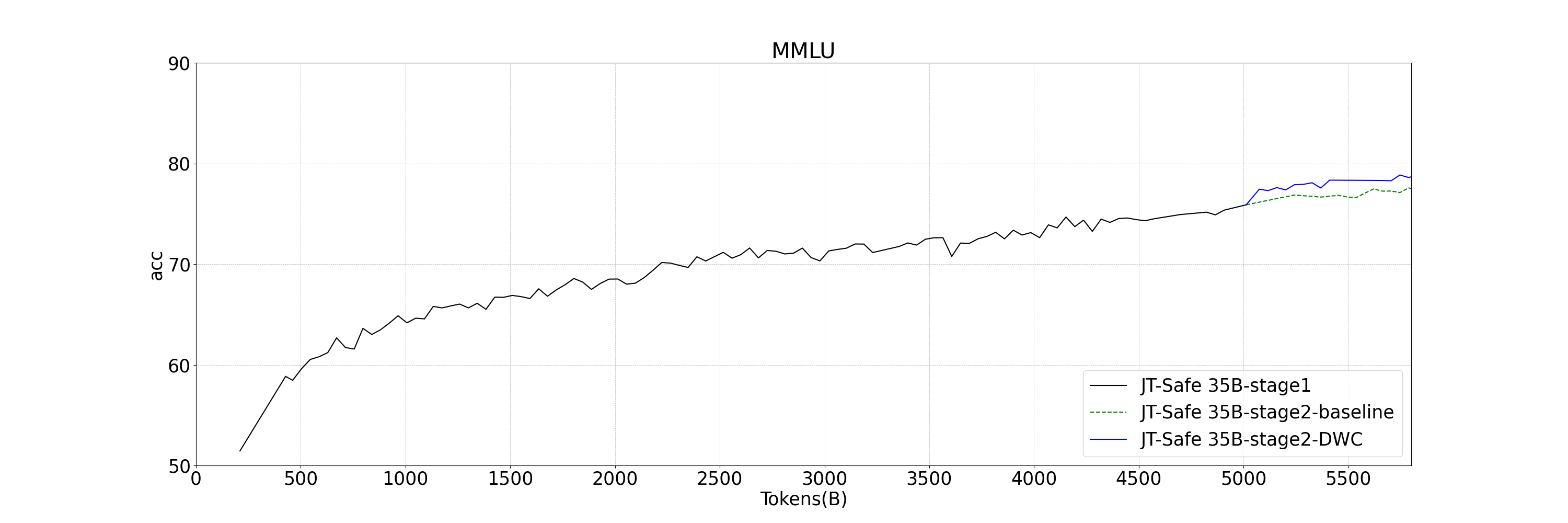} % 调整图片宽度以适应您的需求，例如 0.8\columnwidth
	\caption{A comparison of MMLU scores before and after the addition of DWC .}
	\label{fig:A comparison of MMLU scores before and after the addition of DWC}
\end{figure}	

\begin{figure}[b]
	\centering % 这一行让整个 figure 环境中的内容（即您的图片）居中
	\includegraphics[width=0.8\columnwidth]{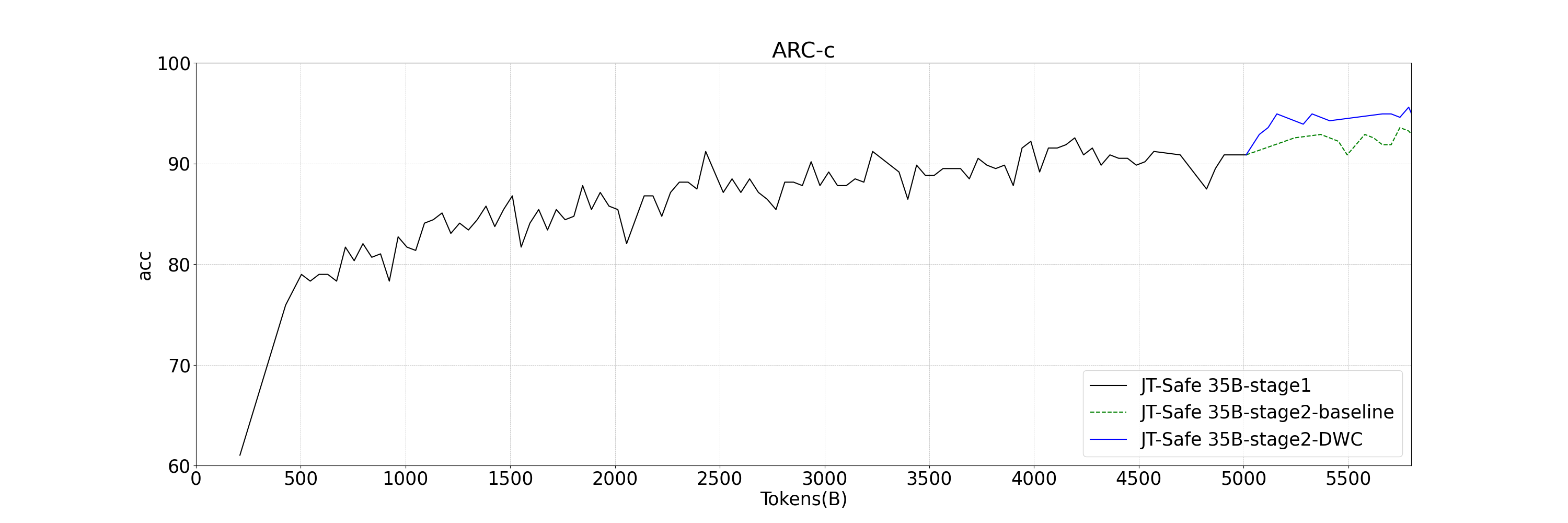} % 调整图片宽度以适应您的需求，例如 0.8\columnwidth
	\caption{A comparison of ARC-c scores before and after the addition of DWC.}
	\label{fig:A comparison of ARC-c scores before and after the addition of DWC}
\end{figure}

\subsection{Benchmarks after Post-training}
Our evaluation framework covers four domains, utilizing benchmarks widely recognized in both academia and industry.
\begin{itemize}
    \item \textbf{General Capabilities}: We select a suite of industry-standard benchmarks to assess the model's core capabilities in high-order reasoning, code generation, knowledge understanding, instruction following, and fundamental mathematics. These include MMLU \citep{hendrycks2020measuring}, MMLU-Pro \citep{wang2024mmlu}, C-Eval \citep{huang2023c}, MBPP \citep{austin2021program}, IFEVAL \citep{wei2024measuring}, GPQA-Diamond \citep{rein2024gpqa}, MATH-500 \citep{lightman2023let},  and AIME-2024 \citep{aime2024}.
    \item \textbf{Industry Capabilities}: We employ our CCR-Bench \citep{xxn2025ccrbench}, a benchmark designed for complex Industry scenarios. We evaluate its two key subtasks, Hard Satisfaction Recall (HSR) and Soft Satisfaction Recall (SSR), to measure the model's proficiency in solving complex real-world problems.
    \item \textbf{Safety Capabilities}: To comprehensively assess the model's safety alignment, we use a set of benchmarks covering multiple risk dimensions, including Flames \citep{huang2023flames}, XSTest \citep{rottger2023xstest}, Forbidden \citep{shen2024anything}, and StrongReject \citep{souly2024strongreject}. These benchmarks measure the model's performance in refusing harmful instructions, avoiding bias and discrimination, and handling sensitive topics.
    \item \textbf{Trustworthiness Capabilities}: We use ChineseSimpleQA \citep{he2024chinese}, proposed by Alibaba, to evaluate the model's hallucination levels and factual accuracy in Chinese open-domain question answering.
\end{itemize}

\begin{table}[t]
\centering
\resizebox{0.89\columnwidth}{!}{%
\begin{tabular}{@{}c|c|c|c|c|>{\columncolor{cyan!12}}c}
\toprule
\textbf{Capabilities}                       & \textbf{Benchmarks} & \textbf{JT-35B-Base} & \textbf{JT-35B} & \textbf{JT-Safe-35B-Base} & \cellcolor{cyan!12} \textbf{JT-Safe-35B} \\ \midrule
\multirow{9}{*}{\textbf{General}}           & MMLU               & 77.37                & 79.53               & 80.07                     & 80.46                \\
                                            & MMLU-Pro           & 54.07                & 60.14               & 49.76                     & 65.12                \\
                                            & C-Eval             & 79.90                & 81.00               & 80.86                     & 81.86                \\
                                            & MBPP               & 59.80                & 68.40               & 62.60                     & 68.40                \\
                                            & IFEVAL             & 54.34                & 86.12               & 64.99                     & 89.28                \\
                                            & GPQA-Diamond      & 37.37                & 50.00               & 41.92                     & 66.81                \\
                                            & MATH-500           & 35.60                & 80.10               & 51.80                     & 84.84                \\
                                            % & GSM8K              & 81.05                & 89.39               & 84.76                     & 94.54                \\
                                            & AIME-2024          & 20.00                & 23.33               & 20.00                     & 36.67                \\ \midrule
\multirow{2}{*}{\textbf{Industry}} & CCR-Bench (HSR)    & 13.20                & 26.20               & 15.80                     & 31.70                \\
                                            & CCR-Bench (SSR)    & 22.01                & 71.40               & 22.46                     & 72.20                \\ \midrule
\multirow{4}{*}{\textbf{Safety}}            & Flames             & 80.70                & 97.98               & 84.35                     & 99.38                \\
                                            & XSTest             & 84.19                & 93.33               & 86.59                     & 94.76                \\
                                            & Forbidden          & 82.74                & 99.13               & 87.03                     & 99.44                \\
                                            & StrongReject       & 84.26                & 92.05               & 84.20                     & 95.71                \\ \midrule
\textbf{Trustworthiness}                        & ChineseSimpleQA    & 29.10                & 59.30               & 29.70                     & 69.80                \\ \bottomrule
\end{tabular}
}
\caption{Evaluation results of the JT-Safe-35B series models across four dimensions.}
\label{analysis_results}
\end{table}

\subsection{Analysis}
We evaluate the four models across the general, industry, safety, and trustworthiness dimensions, with the results presented in Table \ref{analysis_results}. The findings reveal a powerful synergy between the DWC data and our post-training techniques, which validates the effectiveness of our methodology.
A comparison between JT-Safe-35B-Base and JT-35B-Base demonstrates that incorporating DWC data during the pre-training stage effectively enhances the model's inherent safety and trustworthy capabilities. The former achieves superior performance on nearly all safety metrics; for instance, its refusal rate on Flames increases from 80.70\% to 84.35\%, and on Forbidden from 82.74\% to 87.03\%. This indicates that DWC successfully establishes a robust safety foundation from the outset.

Furthermore, the positive foundation established by DWC in pre-training is not only maintained but also amplified during the post-training phase. After post-training, JT-Safe-35B comprehensively surpasses JT-35B in both safety and trustworthiness. On the challenging adversarial benchmark StrongReject, JT-Safe-35B achieves a refusal rate of 95.71\%, significantly higher than the 92.05\% of JT-35B. Similarly, its score on the ChineseSimpleQA trustworthiness test improves from 59.30\% to 69.80\%. This is attributable to our proposed reinforcement learning algorithm, which effectively leverages the DWC data to elevate the model's safety and trustworthiness further.

Finally, the combination of DWC and post-training techniques yields significant synergistic gains across all capability dimensions. The final JT-Safe-35B model achieves the best performance on all general and Industry benchmarks, outperforming all intermediate variants. Notably, on high-order reasoning tasks, its GPQA-Diamond score reaches 66.81\%, far exceeding the 50.00\% of JT-35B, and its 36.67\% score on the AIME-2024 mathematics competition also shows a substantial lead. This suggests that DWC, by providing a higher-quality and more principled data distribution, lays the groundwork for the post-training phase to unlock the model's deeper potential in advanced reasoning, mathematics, and industry-specific applications.

\subsection{Comparison with Baselines}
This section evaluates the performance of the JT-Safe-35B model against state-of-the-art models on established benchmarks. The core objective of this experimental design is to validate the guiding principle of JT-Safe-35B: achieve superior safety performance while maintaining state-of-the-art general capabilities.

\textbf{Baseline Models.} To ensure a fair and comprehensive evaluation, we select the following three representative, state-of-the-art large models as baselines:
\begin{itemize}
    \item \textbf{Qwen3-235B-A22B Non-Thinking} \citep{yang2025qwen3}: This model represents the flagship architecture of the Qwen3 series, featuring a MoE design with a total parameter count of 235 billion, of which only 22 billion are activated per token through a sparse activation mechanism. Qwen3-235B-A22B supports Thinking mode and Non-Thinking mode. In this paper, we use the Non-Thinking mode.
    \item \textbf{Qwen3-30B-A3B-Instruct-2507} \citep{yang2025qwen3}: This lightweight model variant in the Qwen3 series employs a MoE architecture with a total parameter count of 30 billion, activating only 3 billion parameters per token. Optimized for Non-Thinking mode operation, it provides a context length of up to 256K tokens and delivers rapid response times for general-purpose tasks. 
    \item \textbf{DeepSeek-V3.1} \citep{liu2024deepseek}: A model that specializes in code and reasoning. Its architecture and training data are specifically designed to enhance performance on complex logical reasoning and programming tasks. It excels on multiple technical benchmarks and serves as a key reference for evaluating high-order reasoning capabilities.
\end{itemize}

\begin{table}[h]
\centering
\begin{tabular}{@{}l|c|c|c|c|c@{}}
\toprule
\textbf{Models}           & \textbf{General} & \textbf{Industry} & \textbf{Safety} & \textbf{Trustworthy} & \textbf{AVG} \\ \midrule
\textbf{Qwen3-235B-A22B Non Thinking} & 72.85            & 40.60             & 97.63           & 81.08                & 73.04        \\
\textbf{Qwen3-30B-A3B-Instruct-2507}       & 70.52            & 30.60             & 95.62           & 67.92                & 66.17        \\
\textbf{DeepSeek-V3.1}   & 77.93            & 33.50             & 93.98           & 67.32                & 68.18        \\ 
\textbf{JT-Safe-35B}     & 71.68            & 51.95             & 97.32           & 69.80                & 72.69        \\ \bottomrule
\end{tabular}
\caption{A comparative overview of the comprehensive performance of JT-Safe-35B and other models.}
\label{comparative_exp}
\end{table}

The comparison results are shown in Table \ref{comparative_exp}. In terms of general capabilities, JT-Safe-35B obtains a high score of 71.68\%, surpassing the similarly-sized Qwen3-30B-A3B-Instruct-2507 (70.52\%). It remains competitive with the reasoning-specialized DeepSeek-V3.1 (77.93\%). This strongly demonstrates that our model undergoes safety alignment without sacrificing core cognitive abilities. In terms of safety, JT-Safe-35B, with a score of 97.32\%, is on par with the top-tier Qwen3-235B-A22B Non Thinking and significantly outperforms DeepSeek-V3.1 (93.98\%), highlighting the remarkable effectiveness of our safety alignment techniques.

Most notably, in Industry capabilities, JT-Safe-35B holds a commanding lead with a score of 51.95\%, far surpassing all baseline models. This fully validates its immense potential in solving complex real-world problems. Although the Qwen series models exhibit stronger performance in the single dimension of trustworthiness, JT-Safe-35B achieves an good performance balance overall. It successfully avoids the common pitfall of sacrificing performance in other areas in pursuit of a single capability, indicating that our technical method produces a model that is capable, safe, reliable, and highly practical.

In summary, the JT-Safe-35B model lays a foundation of safety and trustworthiness during pre-training and unlocks the model's full potential during post-training, successfully constructing a language model that achieves good performance across general capabilities, industry applications, safety, and trustworthiness.

% \section{the rest is from JTcode}
\section{Conclusion}

This report addresses the fundamental challenges of safety and trustworthiness in current LLMs, and we propose a new pre-training paradigm that builds these capabilities endogenously from the data source. Our core contribution is the innovative integration of "Full-Context Information"—metadata such as source and safety ratings—into our 6.2T token high-quality corpus, alongside a professional vertical domain knowledge base. This approach embeds a traceable and controllable foundation in the model during pretraining. Critically, in the post-training phase, we leverage Reinforcement Learning to activate this embedded contextual information, creating a synergy with alignment training that significantly enhances the model's trustworthy behaviors. Our research confirms that this systematic, data-centric methodology transcends traditional post hoc controls, offering a fundamental pathway toward building genuinely safe, trustworthy, and industry-applicable LLMs.

\bibliographystyle{unsrtnat}
\bibliography{references}  %%% Uncomment this line and comment out the ``thebibliography'' section below to use the external .bib file (using bibtex) .

\begin{thebibliography}{43}
\providecommand{\natexlab}[1]{#1}
\providecommand{\url}[1]{\texttt{#1}}
\expandafter\ifx\csname urlstyle\endcsname\relax
  \providecommand{\doi}[1]{doi: #1}\else
  \providecommand{\doi}{doi: \begingroup \urlstyle{rm}\Url}\fi

\bibitem[Qwen(2025)]{qwen3guard}
Qwen.
\newblock Qwen3guard technical report, 2025.
\newblock URL \url{http://arxiv.org/abs/2510.14276}.

\bibitem[Kalai et~al.(2025)Kalai, Nachum, Vempala, and Zhang]{kalai2025language}
Adam~Tauman Kalai, Ofir Nachum, Santosh~S Vempala, and Edwin Zhang.
\newblock Why language models hallucinate.
\newblock \emph{arXiv preprint arXiv:2509.04664}, 2025.

\bibitem[Bai et~al.(2022{\natexlab{a}})Bai, Kadavath, Kundu, Askell, Kernion, Jones, Chen, Goldie, Mirhoseini, McKinnon, Chen, Olsson, Olah, Hernandez, Drain, Ganguli, Li, Tran-Johnson, Perez, Kerr, Mueller, Ladish, Landau, Ndousse, Lukosuite, Lovitt, Sellitto, Elhage, Schiefer, Mercado, DasSarma, Lasenby, Larson, Ringer, Johnston, Kravec, Showk, Fort, Lanham, Telleen-Lawton, Conerly, Henighan, Hume, Bowman, Hatfield-Dodds, Mann, Amodei, Joseph, McCandlish, Brown, and Kaplan]{bai2022constitutionalaiharmlessnessai}
Yuntao Bai, Saurav Kadavath, Sandipan Kundu, Amanda Askell, Jackson Kernion, Andy Jones, Anna Chen, Anna Goldie, Azalia Mirhoseini, Cameron McKinnon, Carol Chen, Catherine Olsson, Christopher Olah, Danny Hernandez, Dawn Drain, Deep Ganguli, Dustin Li, Eli Tran-Johnson, Ethan Perez, Jamie Kerr, Jared Mueller, Jeffrey Ladish, Joshua Landau, Kamal Ndousse, Kamile Lukosuite, Liane Lovitt, Michael Sellitto, Nelson Elhage, Nicholas Schiefer, Noemi Mercado, Nova DasSarma, Robert Lasenby, Robin Larson, Sam Ringer, Scott Johnston, Shauna Kravec, Sheer~El Showk, Stanislav Fort, Tamera Lanham, Timothy Telleen-Lawton, Tom Conerly, Tom Henighan, Tristan Hume, Samuel~R. Bowman, Zac Hatfield-Dodds, Ben Mann, Dario Amodei, Nicholas Joseph, Sam McCandlish, Tom Brown, and Jared Kaplan.
\newblock Constitutional ai: Harmlessness from ai feedback, 2022{\natexlab{a}}.
\newblock URL \url{https://arxiv.org/abs/2212.08073}.

\bibitem[Inan et~al.(2023)Inan, Upasani, Chi, Rungta, Iyer, Mao, Tontchev, Hu, Fuller, Testuggine, et~al.]{inan2023llama}
Hakan Inan, Kartikeya Upasani, Jianfeng Chi, Rashi Rungta, Krithika Iyer, Yuning Mao, Michael Tontchev, Qing Hu, Brian Fuller, Davide Testuggine, et~al.
\newblock Llama guard: Llm-based input-output safeguard for human-ai conversations.
\newblock \emph{arXiv preprint arXiv:2312.06674}, 2023.

\bibitem[Han et~al.(2024)Han, Rao, Ettinger, Jiang, Lin, Lambert, Choi, and Dziri]{han2024wildguard}
Seungju Han, Kavel Rao, Allyson Ettinger, Liwei Jiang, Bill~Yuchen Lin, Nathan Lambert, Yejin Choi, and Nouha Dziri.
\newblock Wildguard: Open one-stop moderation tools for safety risks, jailbreaks, and refusals of llms.
\newblock \emph{Advances in Neural Information Processing Systems}, 37:\penalty0 8093--8131, 2024.

\bibitem[{OpenAI}(2025)]{openai_gpt5_2025}
{OpenAI}.
\newblock Introducing gpt-5.
\newblock \url{https://openai.com/blog/introducing-gpt-5}, August 2025.
\newblock Accessed: 2024-10-27.

\bibitem[Ouyang et~al.(2022)Ouyang, Wu, Jiang, Almeida, Wainwright, Mishkin, Zhang, Agarwal, Slama, Ray, et~al.]{ouyang2022training}
Long Ouyang, Jeffrey Wu, Xu~Jiang, Diogo Almeida, Carroll Wainwright, Pamela Mishkin, Chong Zhang, Sandhini Agarwal, Katarina Slama, Alex Ray, et~al.
\newblock Training language models to follow instructions with human feedback.
\newblock \emph{Advances in neural information processing systems}, 35:\penalty0 27730--27744, 2022.

\bibitem[Lee et~al.(2023)Lee, Phatale, Mansoor, Mesnard, Ferret, Lu, Bishop, Hall, Carbune, Rastogi, et~al.]{lee2023rlaif}
Harrison Lee, Samrat Phatale, Hassan Mansoor, Thomas Mesnard, Johan Ferret, Kellie Lu, Colton Bishop, Ethan Hall, Victor Carbune, Abhinav Rastogi, et~al.
\newblock Rlaif vs. rlhf: Scaling reinforcement learning from human feedback with ai feedback.
\newblock \emph{arXiv preprint arXiv:2309.00267}, 2023.

\bibitem[Lin et~al.(2025)Lin, Ghosh, Low, Shrivastava, and Mohan]{lin2025refrag}
Xiaoqiang Lin, Aritra Ghosh, Bryan Kian~Hsiang Low, Anshumali Shrivastava, and Vijai Mohan.
\newblock Refrag: Rethinking rag based decoding.
\newblock \emph{arXiv preprint arXiv:2509.01092}, 2025.

\bibitem[Liu et~al.(2025)Liu, Han, Deng, and Feng]{jiutianliujie2025}
Jie Liu, Xue Han, Chao Deng, and Junlan Feng.
\newblock Improving self-consistency for open-domain question answering via automatic prompt engineering and ensemble learning.
\newblock In \emph{Natural Language Processing and Chinese Computing}, 2025.

\bibitem[Weng(2025)]{weng2025think}
Lilian Weng.
\newblock Why we think.
\newblock \emph{lilianweng.github.io}, May 2025.
\newblock URL \url{https://lilianweng.github.io/posts/2025-05-01-thinking/}.

\bibitem[Yang et~al.(2025)Yang, Li, Yang, Zhang, Hui, Zheng, Yu, Gao, Huang, Lv, et~al.]{yang2025qwen3}
An~Yang, Anfeng Li, Baosong Yang, Beichen Zhang, Binyuan Hui, Bo~Zheng, Bowen Yu, Chang Gao, Chengen Huang, Chenxu Lv, et~al.
\newblock Qwen3 technical report.
\newblock \emph{arXiv preprint arXiv:2505.09388}, 2025.

\bibitem[Hatamizadeh et~al.(2025)Hatamizadeh, Akter, Prabhumoye, Kautz, Patwary, Shoeybi, Catanzaro, and Choi]{hatamizadeh2025rlpreinforcementpretrainingobjective}
Ali Hatamizadeh, Syeda~Nahida Akter, Shrimai Prabhumoye, Jan Kautz, Mostofa Patwary, Mohammad Shoeybi, Bryan Catanzaro, and Yejin Choi.
\newblock Rlp: Reinforcement as a pretraining objective, 2025.
\newblock URL \url{https://arxiv.org/abs/2510.01265}.

\bibitem[Erk and Smith(2016)]{erk2016acl}
K.~Erk and Noah~A. Smith.
\newblock Proceedings of the 54th annual meeting of the association for computational linguistics (acl).
\newblock In \emph{Neural Machine Translation of Rare Words with Subword Units}, 2016.

\bibitem[Ding et~al.(2024)Ding, Wang, Paolini, Kumar, Deoras, Roth, and Soatto]{hantian2025icml}
Hantian Ding, Zijian Wang, Giovanni Paolini, Varun Kumar, Anoop Deoras, Dan Roth, and Stefano Soatto.
\newblock Proceedings of the 41st international conference on machine learning (icml 2024).
\newblock In \emph{Fewer truncations improve language modeling}, 2024.

\bibitem[{HUAWEI}(2024)]{huawei910b}
{HUAWEI}.
\newblock Introducing huawei altas.
\newblock \url{https://www.hiascend.com/zh/hardware/product}, October 2024.
\newblock Accessed: 2024-10-17.

\bibitem[{Ainslie} et~al.(2023){Ainslie}, {Lee-Thorp}, {de Jong}, {Zemlyanskiy}, {Lebr{\'o}n}, and {Sanghai}]{Joshua2023GQA}
Joshua {Ainslie}, James {Lee-Thorp}, Michiel {de Jong}, Yury {Zemlyanskiy}, Federico {Lebr{\'o}n}, and Sumit {Sanghai}.
\newblock {GQA: Training Generalized Multi-Query Transformer Models from Multi-Head Checkpoints}.
\newblock \emph{arXiv e-prints}, art. arXiv:2305.13245, May 2023.
\newblock \doi{10.48550/arXiv.2305.13245}.

\bibitem[{Shazeer}(2020)]{Sha2020SiwGLU}
Noam {Shazeer}.
\newblock {GLU Variants Improve Transformer}.
\newblock \emph{arXiv e-prints}, art. arXiv:2002.05202, February 2020.
\newblock \doi{10.48550/arXiv.2002.05202}.

\bibitem[{Su} et~al.(2021){Su}, {Lu}, {Pan}, {Murtadha}, {Wen}, and {Liu}]{su2021RoPE}
Jianlin {Su}, Yu~{Lu}, Shengfeng {Pan}, Ahmed {Murtadha}, Bo~{Wen}, and Yunfeng {Liu}.
\newblock {RoFormer: Enhanced Transformer with Rotary Position Embedding}.
\newblock \emph{arXiv e-prints}, art. arXiv:2104.09864, April 2021.
\newblock \doi{10.48550/arXiv.2104.09864}.

\bibitem[{Zhang} and {Sennrich}(2019)]{zhang2019RMSnorm}
Biao {Zhang} and Rico {Sennrich}.
\newblock {Root Mean Square Layer Normalization}.
\newblock \emph{arXiv e-prints}, art. arXiv:1910.07467, October 2019.
\newblock \doi{10.48550/arXiv.1910.07467}.

\bibitem[Li et~al.(2025)Li, Ma, Yan, Zhang, Liu, Lu, Xu, Chen, Wang, Zhan, et~al.]{li2025model}
Yunshui Li, Yiyuan Ma, Shen Yan, Chaoyi Zhang, Jing Liu, Jianqiao Lu, Ziwen Xu, Mengzhao Chen, Minrui Wang, Shiyi Zhan, et~al.
\newblock Model merging in pre-training of large language models.
\newblock \emph{arXiv preprint arXiv:2505.12082}, 2025.

\bibitem[Si et~al.(2025)Si, Lv, Jiang, Wang, Wang, Yang, Su, Zheng, and Shen]{si2025nan}
Chongjie Si, Kangtao Lv, Jingjing Jiang, Yadao Wang, Yongwei Wang, Xiaokang Yang, Wenbo Su, Bo~Zheng, and Wei Shen.
\newblock Nan: A training-free solution to coefficient estimation in model merging.
\newblock \emph{arXiv preprint arXiv:2505.16148}, 2025.

\bibitem[Li et~al.(2024)Li, Chiang, Frick, Dunlap, Wu, Zhu, Gonzalez, and Stoica]{li2024crowdsourced}
Tianle Li, Wei-Lin Chiang, Evan Frick, Lisa Dunlap, Tianhao Wu, Banghua Zhu, Joseph~E Gonzalez, and Ion Stoica.
\newblock From crowdsourced data to high-quality benchmarks: Arena-hard and benchbuilder pipeline.
\newblock \emph{arXiv preprint arXiv:2406.11939}, 2024.

\bibitem[Wu et~al.(2025)Wu, Mei, Yan, Li, Lai, Ren, Wang, Zhang, Wu, Jin, et~al.]{writingbench2025}
Yuning Wu, Jiahao Mei, Ming Yan, Chenliang Li, Shaopeng Lai, Yuran Ren, Zijia Wang, Ji~Zhang, Mengyue Wu, Qin Jin, et~al.
\newblock Writingbench: A comprehensive benchmark for generative writing.
\newblock \emph{arXiv preprint arXiv:2503.05244}, 2025.

\bibitem[Bai et~al.(2022{\natexlab{b}})Bai, Kadavath, Kundu, Askell, Kernion, Jones, Chen, Goldie, Mirhoseini, McKinnon, et~al.]{cai2022}
Yuntao Bai, Saurav Kadavath, Sandipan Kundu, Amanda Askell, Jackson Kernion, Andy Jones, Anna Chen, Anna Goldie, Azalia Mirhoseini, Cameron McKinnon, et~al.
\newblock Constitutional ai: Harmlessness from ai feedback.
\newblock \emph{arXiv preprint arXiv:2212.08073}, 2022{\natexlab{b}}.

\bibitem[Shao et~al.(2024)Shao, Wang, Zhu, Xu, Song, Bi, Zhang, Zhang, Li, Wu, et~al.]{shao2024deepseekmath}
Zhihong Shao, Peiyi Wang, Qihao Zhu, Runxin Xu, Junxiao Song, Xiao Bi, Haowei Zhang, Mingchuan Zhang, YK~Li, Yang Wu, et~al.
\newblock Deepseekmath: Pushing the limits of mathematical reasoning in open language models.
\newblock \emph{arXiv preprint arXiv:2402.03300}, 2024.

\bibitem[Hendrycks et~al.(2020)Hendrycks, Burns, Basart, Zou, Mazeika, Song, and Steinhardt]{hendrycks2020measuring}
Dan Hendrycks, Collin Burns, Steven Basart, Andy Zou, Mantas Mazeika, Dawn Song, and Jacob Steinhardt.
\newblock Measuring massive multitask language understanding.
\newblock \emph{arXiv preprint arXiv:2009.03300}, 2020.

\bibitem[Huang et~al.(2023{\natexlab{a}})Huang, Bai, Zhu, Zhang, Zhang, Su, Liu, Lv, Zhang, Fu, et~al.]{huang2023c}
Yuzhen Huang, Yuzhuo Bai, Zhihao Zhu, Junlei Zhang, Jinghan Zhang, Tangjun Su, Junteng Liu, Chuancheng Lv, Yikai Zhang, Yao Fu, et~al.
\newblock C-eval: A multi-level multi-discipline chinese evaluation suite for foundation models.
\newblock \emph{Advances in Neural Information Processing Systems}, 36:\penalty0 62991--63010, 2023{\natexlab{a}}.

\bibitem[Cobbe et~al.(2021)Cobbe, Kosaraju, Bavarian, Chen, Jun, Kaiser, Plappert, Tworek, Hilton, Nakano, et~al.]{cobbe2021training}
Karl Cobbe, Vineet Kosaraju, Mohammad Bavarian, Mark Chen, Heewoo Jun, Lukasz Kaiser, Matthias Plappert, Jerry Tworek, Jacob Hilton, Reiichiro Nakano, et~al.
\newblock Training verifiers to solve math word problems.
\newblock \emph{arXiv preprint arXiv:2110.14168}, 2021.

\bibitem[Lightman et~al.(2023)Lightman, Kosaraju, Burda, Edwards, Baker, Lee, Leike, Schulman, Sutskever, and Cobbe]{lightman2023let}
Hunter Lightman, Vineet Kosaraju, Yuri Burda, Harrison Edwards, Bowen Baker, Teddy Lee, Jan Leike, John Schulman, Ilya Sutskever, and Karl Cobbe.
\newblock Let's verify step by step.
\newblock In \emph{The Twelfth International Conference on Learning Representations}, 2023.

\bibitem[Chen et~al.(2021)Chen, Tworek, Jun, Yuan, Pinto, Kaplan, Edwards, Burda, Joseph, Brockman, et~al.]{chen2021evaluating}
Mark Chen, Jerry Tworek, Heewoo Jun, Qiming Yuan, Henrique Ponde De~Oliveira Pinto, Jared Kaplan, Harri Edwards, Yuri Burda, Nicholas Joseph, Greg Brockman, et~al.
\newblock Evaluating large language models trained on code.
\newblock \emph{arXiv preprint arXiv:2107.03374}, 2021.

\bibitem[Wei et~al.(2024)Wei, Karina, Chung, Jiao, Papay, Glaese, Schulman, and Fedus]{wei2024measuring}
Jason Wei, Nguyen Karina, Hyung~Won Chung, Yunxin~Joy Jiao, Spencer Papay, Amelia Glaese, John Schulman, and William Fedus.
\newblock Measuring short-form factuality in large language models.
\newblock \emph{arXiv preprint arXiv:2411.04368}, 2024.

\bibitem[Wang et~al.(2024)Wang, Ma, Zhang, Ni, Chandra, Guo, Ren, Arulraj, He, Jiang, et~al.]{wang2024mmlu}
Yubo Wang, Xueguang Ma, Ge~Zhang, Yuansheng Ni, Abhranil Chandra, Shiguang Guo, Weiming Ren, Aaran Arulraj, Xuan He, Ziyan Jiang, et~al.
\newblock Mmlu-pro: A more robust and challenging multi-task language understanding benchmark.
\newblock \emph{Advances in Neural Information Processing Systems}, 37:\penalty0 95266--95290, 2024.

\bibitem[Austin et~al.(2021)Austin, Odena, Nye, Bosma, Michalewski, Dohan, Jiang, Cai, Terry, Le, et~al.]{austin2021program}
Jacob Austin, Augustus Odena, Maxwell Nye, Maarten Bosma, Henryk Michalewski, David Dohan, Ellen Jiang, Carrie Cai, Michael Terry, Quoc Le, et~al.
\newblock Program synthesis with large language models.
\newblock \emph{arXiv preprint arXiv:2108.07732}, 2021.

\bibitem[Rein et~al.(2024)Rein, Hou, Stickland, Petty, Pang, Dirani, Michael, and Bowman]{rein2024gpqa}
David Rein, Betty~Li Hou, Asa~Cooper Stickland, Jackson Petty, Richard~Yuanzhe Pang, Julien Dirani, Julian Michael, and Samuel~R Bowman.
\newblock Gpqa: A graduate-level google-proof q\&a benchmark.
\newblock In \emph{First Conference on Language Modeling}, 2024.

\bibitem[AIME(February 2024)]{aime2024}
AIME.
\newblock American invitational mathematics examination - aime.
\newblock \emph{In American Invitational Mathematics Examination - AIME 2024}, February 2024.

\bibitem[Xiaona et~al.(2025)Xiaona, Yiqiao, Jiacheng, Yuanhang, Huiqi, Yunfei, Xinbao, Minglu, Fanyu, Chao, and Feng]{xxn2025ccrbench}
Xue Xiaona, Huang Yiqiao, Li~Jiacheng, Zheng Yuanhang, Miao Huiqi, Ma~Yunfei, Sun Xinbao, Liu Minglu, Meng Fanyu, Deng Chao, and Junlan Feng.
\newblock Ccr-bench: A comprehensive benchmark for evaluating llms on complex constraints, control flows, and real-world cases.
\newblock \emph{https://huggingface.co/datasets/JT-LM/CCR-Bench}, 2025.

\bibitem[Huang et~al.(2023{\natexlab{b}})Huang, Liu, Guo, Sun, Sun, Wang, Zhou, Wang, Teng, Qiu, et~al.]{huang2023flames}
Kexin Huang, Xiangyang Liu, Qianyu Guo, Tianxiang Sun, Jiawei Sun, Yaru Wang, Zeyang Zhou, Yixu Wang, Yan Teng, Xipeng Qiu, et~al.
\newblock Flames: Benchmarking value alignment of llms in chinese.
\newblock \emph{arXiv preprint arXiv:2311.06899}, 2023{\natexlab{b}}.

\bibitem[R{\"o}ttger et~al.(2023)R{\"o}ttger, Kirk, Vidgen, Attanasio, Bianchi, and Hovy]{rottger2023xstest}
Paul R{\"o}ttger, Hannah~Rose Kirk, Bertie Vidgen, Giuseppe Attanasio, Federico Bianchi, and Dirk Hovy.
\newblock Xstest: A test suite for identifying exaggerated safety behaviours in large language models.
\newblock \emph{arXiv preprint arXiv:2308.01263}, 2023.

\bibitem[Shen et~al.(2024)Shen, Chen, Backes, Shen, and Zhang]{shen2024anything}
Xinyue Shen, Zeyuan Chen, Michael Backes, Yun Shen, and Yang Zhang.
\newblock " do anything now": Characterizing and evaluating in-the-wild jailbreak prompts on large language models.
\newblock In \emph{Proceedings of the 2024 on ACM SIGSAC Conference on Computer and Communications Security}, pages 1671--1685, 2024.

\bibitem[Souly et~al.(2024)Souly, Lu, Bowen, Trinh, Hsieh, Pandey, Abbeel, Svegliato, Emmons, Watkins, et~al.]{souly2024strongreject}
Alexandra Souly, Qingyuan Lu, Dillon Bowen, Tu~Trinh, Elvis Hsieh, Sana Pandey, Pieter Abbeel, Justin Svegliato, Scott Emmons, Olivia Watkins, et~al.
\newblock A strongreject for empty jailbreaks.
\newblock \emph{Advances in Neural Information Processing Systems}, 37:\penalty0 125416--125440, 2024.

\bibitem[He et~al.(2024)He, Li, Liu, Tan, Wang, Huang, Bu, Guo, Hu, Zheng, et~al.]{he2024chinese}
Yancheng He, Shilong Li, Jiaheng Liu, Yingshui Tan, Weixun Wang, Hui Huang, Xingyuan Bu, Hangyu Guo, Chengwei Hu, Boren Zheng, et~al.
\newblock Chinese simpleqa: A chinese factuality evaluation for large language models.
\newblock \emph{arXiv preprint arXiv:2411.07140}, 2024.

\bibitem[Liu et~al.(2024)Liu, Feng, Xue, Wang, Wu, Lu, Zhao, Deng, Zhang, Ruan, et~al.]{liu2024deepseek}
Aixin Liu, Bei Feng, Bing Xue, Bingxuan Wang, Bochao Wu, Chengda Lu, Chenggang Zhao, Chengqi Deng, Chenyu Zhang, Chong Ruan, et~al.
\newblock Deepseek-v3 technical report.
\newblock \emph{arXiv preprint arXiv:2412.19437}, 2024.

\end{thebibliography}

\clearpage

\end{document}